# BIGNet: Pretrained Graph Neural Network for Embedding Semantic, Spatial, and Topological Data in BIM Models

Jin Han[1]  |  Xin-Zheng Lu[1]  |  Jia-Rui Lin[1,2,*]

[1]Department of Civil Engineering, Tsinghua University, China

[2]Key Laboratory of Digital Construction and Digital Twin, Ministry of Housing and Urban-Rural Development, China

**Correspondence**
Jia-Rui Lin, Department of Civil Engineering, Tsinghua University, Beijing, China
Email: lin611@tsinghua.edu.cn

**Funding information**
National Natural Science Foundation of China (No. 52378306, No. 52238011)

**ABSTRACT**

Large Foundation Models (LFMs) have demonstrated significant advantages in civil engineering, but they primarily focus on textual and visual data, overlooking the rich semantic, spatial, and topological features in BIM (Building Information Modelling) models. Therefore, this study develops the first large-scale graph neural network (GNN), BIGNet, to learn, and reuse multidimensional design features embedded in BIM models. Firstly, a scalable graph representation is introduced to encode the "semantic-spatial-topological" features of BIM components, and a dataset with nearly 1 million nodes and 3.5 million edges is created. Subsequently, BIGNet is proposed by introducing a new message-passing mechanism to GraphMAE2 and further pretrained with a node masking strategy. Finally, BIGNet is evaluated in various transfer learning tasks for BIM-based design checking. Results show that: 1) homogeneous graph representation outperforms heterogeneous graph in learning design features, 2) considering local spatial relationships in a 30 cm radius enhances performance, and 3) BIGNet with GAT (Graph Attention Network)-based feature extraction achieves the best transfer learning results. This innovation leads to a 72.7% improvement in Average F1-score over non-pretrained models, demonstrating its effectiveness in learning and transferring BIM design features and facilitating their automated application in future design and lifecycle management.

## 1 | INTRODUCTION

Building Information Modeling (BIM) is a digital approach used in construction to create, manage, and optimize detailed 3D models of buildings throughout their entire lifecycle (Hu, Y. et al., 2022). By integrating comprehensive data on a facility's physical and functional characteristics into a single, cohesive model, BIM has brought significant convenience to engineering applications (Hu, Z. et al., 2022). As a result, BIM has been extensively researched in areas such as visualization (Leng et al., 2021), human-machine collaborative design (Lin et al., 2024; Hu et al., 2023), and life cycle assessment (Zheng et al., 2023; Rad et al., 2021), and is rapidly gaining widespread adoption. As a comprehensive digital representation of building characteristics, BIM integrates a broad spectrum of engineering data, including semantic features, spatial layouts, material properties, and other relevant information (He et al., 2025). For instance, historical BIM model data often encapsulates design features derived from engineers' extensive practical experience, such as the standard range of dimensions for doors and windows, as well as spatial and connection relationships between various components. Therefore, extracting, learning, and reusing relevant design knowledge and patterns from this data is very importance to enhance building design, construction, and maintenance.

With the rapid development of artificial intelligence (AI) technologies, powerful capabilities in nonlinear and fuzzy learning from existing data and knowledge have emerged, bringing promising solutions to these challenges and driving intelligent transformation (Han et al., 2024a; Amezquita-Sanchez et al., 2016; Adeli, 2001; Adeli, 1995). However, BIM models typically consist of complex 3D geometric data and non-geometric information, whereas machine learning requires structured tabular data for training (Xiao et al., 2019; Lin et al., 2016; Panakkat & Adeli, 2007). Therefore, a major challenge is how to effectively extract and transform the vast and diverse features of BIM models into formats compatible with machine learning.

Current research is still in the infant phase, primarily focusing on tasks such as clash detection and semantic enrichment. The proposed methods for embedding BIM data can be broadly categorized into two types: 1) Extracting key component properties and converting them into independent tensors for classification tasks or clash detection and prediction (Lin & Huang, 2019; Wang & Leite, 2013; Hu & Castro-Lacouture, 2019; Liu, H et al., 2024; Utkucu et al., 2024; Jang & Lee, 2024). This method overlooks the spatial and topological relationships between components, which are crucial for accurately capturing their interconnections. 2) Representing BIM models as graphs, with nodes and edges denoting component properties and some simple spatial and topological relationships, enabling automated generative design or clash resolution (Wang et al., 2022; Hu et al., 2020; 2023; Li et al., 2024; Kayhani et al., 2023; Liu, X et al., 2024; Liang et al., 2024). While these studies address some limitations of the first method, the spatial and topological relationships they capture are often simplistic and insufficient. Furthermore, the graph construction process is typically tailored to specific tasks, lacking flexibility and general applicability.

Besides the challenges in representing BIM models, the scarcity of labeled data is another key factor limiting the application of deep learning in BIM. However, recent advancements in pretrained and transfer learning models provide a promising solution. These models can leverage large amounts of unlabeled data for self-supervised learning of complex design patterns and expertise, while the acquired general knowledge can be transferred to support various tasks (Weiss et al., 2016; Adeli & Nogay, 2021). Previous studies have demonstrated their significant advantages in knowledge extraction and reuse in the architectural domain, due to their robust transfer learning capabilities (Ge et al., 2024; Han et al., 2024; Zheng et al., 2024). While most pretrained and transfer learning research has focused on textual and visual data, BIM models pose distinct challenges due to their complex integration of semantic, spatial, and topological design features. GNNs provide a potential solution by effectively processing such non-Euclidean spatial data (Garcia & Niepert, 2017; Kipf & Welling, 2016; Cui et al., 2020; Ahmadlou et al., 2021). Inspired by these, this study pre-trains a large-scale GNN to learn design knowledge from BIM models and support multiple downstream tasks.

The design quality of BIM models is critical to the management and operation throughout a building's entire lifecycle. Poor BIM design can cause issues such as construction drawing errors and calculation discrepancies early on and may lead to unforeseen problems during subsequent construction, completion acceptance, and maintenance, ultimately hindering efficient management and sustainable development (Zou et al., 2017). As a result, the checking and refinement of BIM models have gained widespread attention from researchers (Liu, H et al., 2024; Utkucu et al., 2024; Li et al., 2024; Kayhani et al., 2023). However, existing studies mostly focus on clash detection and semantic enrichment, lacking methods applicable to multiple detection tasks. Therefore, this study uses multiple tasks for BIM-based design checking as downstream tasks to validate the effectiveness of the proposed pre-training approach.

In summary, this study introduces BIGNet, the first large-scale pretrained GNN designed to learn, reuse, and transfer the semantic, spatial, and topological features of BIM models for multiple downstream tasks, highlighting that its novelty lies in the pretrained GNN framework rather than the BIM-to-graph conversion itself. First, a scalable graph representation is introduced to encode the "semantic-spatial-topological" design features of BIM components (Section 3.1). Second, BIGNet is proposed by introducing a new message-passing mechanism to GraphMAE2, and further pretrained with a node masking strategy (Section 3.2). Third, BIGNet is evaluated in various transfer learning tasks for BIM-based design checking (Section 3.3). Then, different graph representation methods and transfer learning approaches are investigated and compared with the results obtained without pre-training (Section 4). Finally, Section 5 discusses the contributions of this study and outlines directions for future work, while Section 6 concludes this research.

## 2 | RELATED WORK

### 2.1 | Representation methods of BIM models for deep learning

BIM models represent building information through complex 3D geometric data (e.g., points, lines, surfaces), non-geometric data (e.g., component properties), and relational data (e.g., connection relationships), forming a multi-type, multi-level structure. In contrast, deep neural networks require structured, standardized inputs (e.g., 2D images or tensors), typically numeric and matrix-based, and with fixed dimensions, which differs significantly from BIM's representation (Xiao et al., 2019; Lin et al., 2016; Panakkat & Adeli, 2007). Therefore, it is necessary to extract and transform BIM data into structured formats to suit the requirements of deep learning (DL) models. Existing methods mainly fall into two categories: representing BIM models as tensors or graphs, using machine learning or deep neural networks, and graph neural networks, respectively.

The first approach extracts key component attributes and converts them into independent tensors for training and prediction. Lin & Huang (2019) extracted information from the clash detection report, leaving numeric features (e.g., 'Distance') unchanged, while nominal/text features (e.g., 'ItemType-1') are one-hot encoded. These features are then stacked into tensors and input into machine learning algorithms for clash detection. Wang & Leite (2013) selected three types of attributes: geometric properties (e.g., 'Volume'), usage and material (e.g., 'Material Flexibility'), and clash information (e.g., 'Clashing volume'), which were converted into tensors using the same method, then used to train machine learning algorithms for clash prediction. Liu, H et al. (2024) trained a multi-modal DL model using component images and basic attributes such as thickness and concrete grade to classify components. Wang et al. (2023) employed machine learning algorithms to classify building object geometry images and utilized a rule-based approach for BIM model generation.

Additionally, many other studies have explored the vector-based representation and utilization of BIM models, as summarized in Table 1. Notably, the specific properties extracted for the same attribute may vary across studies. For example, the dimension attribute may include properties such as thickness and length. Therefore, the properties in this table are described in a generalized manner for simplicity, without specific distinctions. Given the strong correlation between property extraction and research objectives, this information is summarized in Table 1.

**TABLE 1** Summary of properties involved in the vectorization of BIM models.

| Type | Property | Example | Lin & Huang, 2019 | Wang & Leite, 2013 | Hu & Castro-Lacouture, 2019 | Liu, H et al., 2024 | Utkucu et al., 2024 | Jang & Lee, 2024 |
|---|---|---|---|---|---|---|---|---|
| Semantic | Shape | Round | | √ | √ | | | |
| | Dimension | 2 (cm) | √ | √ | √ | √ | √ | √ |
| | Volume | 12 (cm2) | | √ | | | | |
| | Identifier | Wall_200 | | | | √ | | |
| | Type | Basic Wall | | | | √ | | |
| | Geometrical property | Euler characteristic is 2 | | | | | √ | |
| | Material | Concrete C35 | | √ | √ | √ | | √ |
| | Structure analysis | Load bearing 500kN | | | | √ | | |
| | Position | Level 2 | √ | | √ | | √ | √ |
| | Design Phase | Early design | | | √ | | | |
| | Related discipline | MEP | | | √ | | | |
| Spatial | Available space | Can be moved upward by 2m | √ | √ | √ | | | |
| Topology | Number of connected components | 2 | | | | | √ | |
| | Application scenarios | | Clashes detection | Clashes resolution | Clashes detection | Component classification | Component classification | Component Matching |

Note: the specific properties extracted for the same attribute may vary across studies. For example, the dimension attribute may include properties such as thickness and length. Therefore, the attributes in this table are described in a general manner for simplicity, without specific distinctions.

While the above methods consider key component properties, their tensor-based representation overlooks critical spatial and topological relationships, such as relative positions, connections, and interactions (e.g., 'force transfer mechanisms'). These relationships are crucial not only for understanding component interactions and functional coordination but also for capturing design patterns. For instance, the spatial relationship between beams and columns reflects the force transfer mechanisms, where beams are typically placed above columns and connected to other beams. Without this information, we cannot fully comprehend how beams and columns share the load.

Therefore, researchers have attempted to represent BIM models as graphs, with components and their relationships as nodes or edges, and properties as node or edge features. For example, Wang et al. (2022) represented point-based instances (e.g., 'Pipe Fitting') and curve-based instances (e.g., 'Pipe Curve') as nodes and edges in a graph, using graph matching to detect connection errors. Hu et al. (2020; 2023) modeled components as nodes and represented spatial and topological relationships (e.g., clashes, impacts, and connections) as edges for collision detection and optimization. Unlike the fine-grained representations above, Li et al. (2024) treated rooms as nodes, with edges representing their hierarchical relationships, enabling the automatic generation of modular building designs. Similarly, Gan (2022) represented rooms as graph nodes and modeled adjacency and connectivity relationships as edges, enabling the automatic generative design of modular buildings. Table 2 summarizes various methods for representing BIM models as graphs, detailing the node, edge, and feature structures, along with their application scenarios. For clarity, node features are generalized, without listing all specific properties. This overview clarifies existing graph construction methods, showing that each approach tailors its structure to specific tasks. While these methods effectively represent certain spatial and topological properties in BIM models, they reveal significant differences in graph construction across tasks, lacking a universal approach to BIM model representation—a crucial prerequisite for broader applicability and integration.

**TABLE 2** Summary of methods for representing BIM models as graphs.

| | Node | Node feature | Edge | Edge feature | Application scenarios |
|---|---|---|---|---|---|
| Wang et al., 2022 | point-based instances | - | curve-based instances | - | Connection logical detection |
| Gan, 2022 | Room | Basic information of components in a room | Adjacency and connections of rooms | - | Generative design of modular buildings |
| Hu et al., 2020; 2023 | Component | Component properties | Clash, connection, and impact relation | Relationship types | Clash resolution |
| Kayhani et al., 2023 | As-designed BIM elements | Semantics, local neighborhood, and PointNet features | Topologically or spatially relation | Relationship types | Automated construction quality assessments |
| Liu, X et al., 2024 | Component | Component properties | Clash, connection, and impact relation | Distance | Clash resolution |
| Liang et al., 2024 | The endpoints of walls and non-overlapping columns | spatial location and environmental information | Walls | Wall types | Automated generative design |
| Li et al., 2024 | Room | Spatial-geometric features of rooms | Connections between rooms | - | Automated generative design |

Note: the papers listed in Tables 1 and 2 were selected based on the following criteria: 1) focus on representing, vectorizing, or learning semantic, spatial, or topological features in BIM models; 2) published in peer-reviewed journals or conferences; and 3) clear relevance to machine learning or graph-based approaches. Meanwhile, representative works covering various modeling methods and application scenarios are included.

## 2.2 | Large foundation model for AEC

Despite the widespread use of BIM applications and the accumulation of numerous building models, there remains a lack of sufficient labeled data. Furthermore, the preparation of adequate training datasets necessitates considerable manual effort, which is both time-consuming and costly (Zheng et al., 2024). For example, the studies mentioned in Section 2.1 typically train DL models on a single dataset, limiting accuracy improvements. Recently, the emergence of pretrained and transfer learning models has provided a new solution to this issue. They leverage self-supervised learning to extract features and contextual information from large amounts of unlabeled data, and then transfer them to specific tasks using a small set of labeled data (Weiss et al., 2016; Adeli & Nogay, 2021). Existing studies have demonstrated significant advantages in knowledge extraction and reuse in the architectural domain, due to their robust transfer learning capabilities (Ge et al., 2024, Zhang et al., 2021). For example, Han et al. (2024) enhanced Weibo data classification accuracy by transferring a BERT model pretrained on a general-domain corpus with minimal labeled data. Zheng et al. (2022) pretrained a large model on an AEC-domain corpus and transferred it to text classification (TC) and named entity recognition (NER) tasks, significantly improving accuracy.

While most research on pretrained and transfer learning has focused on textual and visual data, BIM models present unique challenges due to their complex integration of semantic, spatial, and topological features (Kayhani et al., 2023). Graph neural networks (GNNs) offer a promising solution by effectively processing non-Euclidean spatial data. Self-supervised graph autoencoders (GAEs) can be trained on large unlabeled data by reconstructing input graph data. For instance, EP (Embedding Propagation) recovers vertex features (Garcia & Niepert, 2017), VGAE (Variational Graph Auto-Encoders) reconstructs missing data (Kipf & Welling, 2016), and AGE (Adaptive Graph Encoder) focuses on link prediction and graph clustering (Cui et al., 2020). Building on these studies, Hou et al. (2023) addressed key challenges and significantly improved pretrained large foundation model performance. Thus, this study adopts their GraphMAE2 architecture for pre-training. While the effectiveness of pretrained large foundation graph neural networks has been demonstrated, their application to BIM models, which are highly suited to graph structures, remains unexplored.

## 2.3 | Research gaps and objective

Although some studies have explored the application of DL models to various downstream tasks in BIM by leveraging historical data, there are still three knowledge gaps that this paper aims to address.

(a) Firstly, previous studies of BIM data representation have primarily focused on extracting features tailored specifically to individual tasks. This task-specific focus restricts the generalizability of learned features across diverse applications. Consequently, there is a notable absence of a unified method capable of comprehensively representing and extracting the semantic, spatial, and topological design features inherent in BIM models, thus restricting the large-scale identification and reuse of universal design patterns embedded within historical BIM data.

(b) Secondly, existing studies typically require extensive labeling efforts for each specific task, leading to substantial resource consumption. To date, pretrained models have not been applied to BIM data to effectively learn general design patterns, which would significantly reduce the amount of labeled data required for transfer learning.

(c) Finally, the design quality of BIM models is critical to the lifecycle management and operation of buildings, leading to widespread research on BIM model inspection and refinement. However, existing research primarily addresses only specific, singular error categories and lacks a unified approach capable of simultaneously addressing multiple design checking tasks.

To address the above knowledge gaps, this study aims to propose a BIM-specific scalable graph representation to embed the "semantic-spatial-topological" design features of BIM components. Based on this, a pretrained graph neural network will be developed to extract, learn, and reuse the implicit design features from existing data. Finally, the model will be transferred to multiple BIM-based design checking tasks to validate its effectiveness and generalizability, enabling real-time error detection to ensure smooth project execution at all stages.

## 3 | METHODOLOGY

To achieve these goals, this study proposes the framework in Figure 1. The approach consists of three core steps:

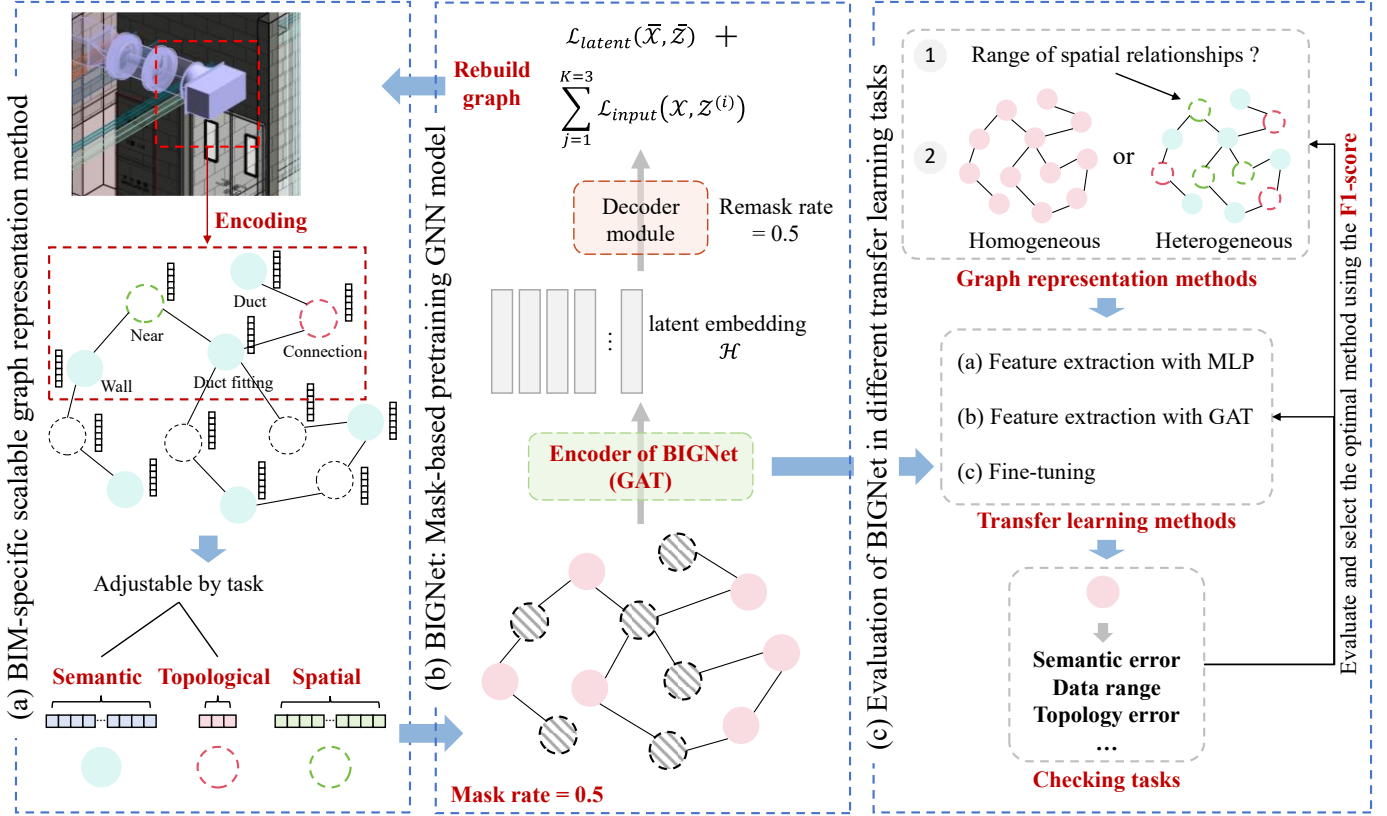

**FIGURE 1** Framework of the proposed method.

Step 1 BIM-specific scalable graph representation method: focuses on achieving a BIM-specific scalable graph representation, embedding the "semantic-spatial-topological" multidimensional features of BIM models (Section 3.1). Unlike previous approaches, which typically focus on task-specific or partial feature extraction, our method achieves a unified and comprehensive encoding of component properties and their spatial/topological relationships. By extending and advancing the unified network-based representation (Han et al., 2025), we develop an end-to-end pipeline for extracting, aligning, and normalizing these features as graph nodes and high-dimensional vector node attributes. This results in the first scalable BIM-specific graph dataset capable of supporting efficient learning and transfer of complex design patterns across diverse downstream tasks, overcoming the fragmentation and limited reusability seen in prior studies.

Step 2 Mask-based pretraining GNN model (BIGNet): aims to pretrain a graph neural network to learn the implicit design patterns and knowledge embedded in BIM data (Section 3.2). After converting the BIM model into a graph, the GraphMAE2 architecture was introduced to pretrain the first graph neural network, BIGNet, which learns and embeds the implicit design features of components. By randomly masking node features, the pretraining task becomes a node feature reconstruction problem, enabling the model to self-supervise and recover the entire graph structure from local information. This enhances the model's understanding of implicit component relationships and provides a stronger feature foundation for transfer learning for small-sample downstream tasks. Notably, our pretrained model is task-agnostic and can be efficiently adapted to multiple distinct BIM downstream tasks within a unified framework.

Step 3 Evaluation of BIGNet in different transfer learning tasks: aims to evaluate the proposed method through multiple tasks for BIM-based design checking (Section 3.3). This study annotated three types of errors from 16 real-world building BIM models as the dataset for transfer learning, including: Semantic conflicts, data range error, and topological error. The design checking performance of the algorithm was evaluated using metrics such as the confusion matrix and F1 score. Based on this dataset and evaluation metrics, the study explores the impact of different graph representations and transfer learning methods on model performance to identify the optimal model, guiding future practical applications. Finally, by comparing with non-pretrained models, the proposed method's applicability and integration are validated.

## 3.1 | A BIM-specific scalable graph representation method

In previous research, a unified network-based representation method for BIM models was proposed (Han et al., 2025). By integrating rich local spatial and topological relationships, this method produces a more informative network, enhancing its suitability for AI learning. Building on this, this study further refines component properties and their relationships through selection, embedding, alignment, and normalization, proposing a BIM-specific scalable graph representation method, as shown in Table 3. Unlike prior approaches that extract features tailored only to specific tasks, the proposed method offers a generalizable representation capable of capturing the diverse information inherent in BIM models. In practice, engineers may select only the attributes and relationships relevant to their specific downstream tasks and encode them within the proposed framework. When computational resources permit, the framework can accommodate a more exhaustive set of features, resulting in a pretrained model with broader applicability across diverse tasks.

The specific data extraction and graph representation methods mainly consist of three parts: Semantic (Section 3.1.1), Topological (Section 3.1.2), and spatial (Section 3.1.3) feature representation of components in the graph.

### 3.1.1 | Semantic feature selection of components

For the downstream tasks of this study, specific component properties in the previously constructed network (Han et al., 2025) were selected to represent semantic features, including shape, dimensions, structural purpose, family, and family type. Specifically, geometric shapes of components are classified into three types: cuboid, cylinder, and irregular shapes. For cuboid components, size properties include length, width, and height; for cylindrical components (e.g., pipes), the cross-sectional radius represents both width and height; and for irregularly shaped components (e.g., certain fittings), three key dimensions representing their geometry are selected. It should be noted that, although shape and dimensions are fundamentally geometric attributes of building components, in this study, they are categorized as semantic features. This classification emphasizes their importance in representing and distinguishing the roles, functions, and intended uses of components within the building's design logic. It also serves to clearly differentiate these intrinsic properties from spatial and topological features, which primarily describe the position and interrelationships among components. To clarify the function of components within a building, their structural purpose—whether structural or non-structural—was also extracted. Additionally, the Family Name and FamilySymbol Name often reflect an object's type, geometric dimensions, material properties, and other detailed attributes. For example, Family Names like "Fixed Window" and "Door," and FamilySymbol Names such as "Architectural Wall_200" and "Exhaust Duct_Galvanized Steel," provide valuable semantic information. Hence, these properties were also extracted as semantic features of the components.

It is important to note that this method is highly versatile and extensible. The proposed approach applies to nearly all text- or numerically-stored features of any component. The properties extracted in this study were selected based on the three BIM-based design checking tasks. For future applications, additional relevant properties can be extracted to explore a broader range of scenarios.

### 3.1.2 | Topological feature selection of components

To represent the topological features of components, this study adopts three topological relationships from previous research: 1) Connection, 2) Touch floor, and 3) Host (Han et al., 2025), as shown in Figure 2 and Table 3. A connection relationship refers to the direct linkage between two components, reflecting mechanical force transfer paths in structural systems or the distribution of fluid or electrical in piping systems. For example, the connection between a duct and its terminal elbow falls into this category. A touch floor relationship describes the vertical connection between components and floor slabs, providing insight into the functional layout and component distribution across different building levels. Examples include the connection between walls and floor slabs or the placement of equipment within a floor. A host relationship represents a component fully embedded within another, aiding in the analysis of functional dependencies between components. An example is the inclusion relationship between doors or windows and walls.

**TABLE 3** Unified embedding for semantic, topological, and spatial features of components.

| | Features | Description | Encoding | Numbers |
|---|---|---|---|---|
| Semantic | shape | Component geometries, such as cubes, cylinders, and other irregular shapes. | One-hot | 3 |
| | Dimensions | length, width, and height of components; for cylinders, both width and height | Numeric | 3 |
| | Component functions | Component structural purpose, categorizing as either structural or non-structural. | One-hot | 2 |
| | Families | Component Family name, describing its function and purpose. | M3E | 64 |
| | FamilySymbol | Component FamilySymbol name, including detailed properties such as specific geometry and material properties. | M3E | 64 |
| Spatial | Positioning coordinates | Coordinates of component positioning points or positioning line endpoints. | Numeric | 6 |
| | Offset | Vertical offset from top or bottom elevation constraint (e.g., top or bottom elevation). | Numeric | 2 |
| | Spatial relationships | Spatial relationships include five categories: 1) different surface, 2) interface non-parallel, 3) interface parallel, 4) point-to-line, and 5) point-to-point. | One-hot | 5 |
| | Angle | Intersection angle between the positioning lines of two components. Note that if one of the components uses point positioning, the angle is 0. | Numeric | 1 |
| | Shortest distance vector coordinates | Endpoint coordinates of the shortest distance vector between the positioning points or lines of two components. | Numeric | 1 |
| | Shortest distance | Actual spatial separation between two components. A negative value indicates an overlap or collision between the components, while a positive value denotes a gap. | Numeric | 3 |
| | Angle with the horizontal plane | For spatial relationships in categories 2-5, the angle with the horizontal plane represents the angle between the plane or line formed by their positioning lines or points and the horizontal plane. In category 1, the angle is 0. | Numeric | 1 |
| Topological | Topological relationships | Topological relationships include three types: 1) Connection: describing direct connections between components, 2) Touch floor: describing vertical connections between components and floors, and 3) Host: describing a component being entirely embedded within another. | One-hot | 3 |

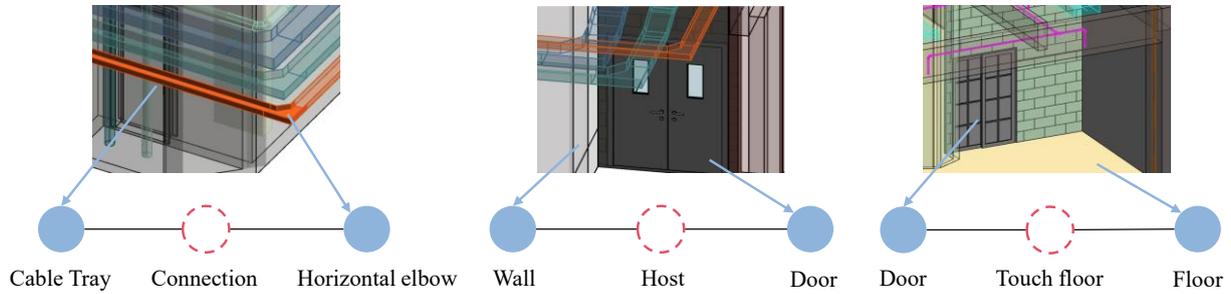

**FIGURE 2** Three types of topological relationships.

### 3.1.3 | Spatial feature selection of components

To represent the spatial features of components, this study includes positioning coordinates and offset properties, which directly indicate component locations. Furthermore, it incorporates spatial relationships from previous research to capture local interactions between components (Han et al., 2025), addressing the challenge of learning complex relative positions from simple coordinate data, as shown in Table 3. Specifically, a spatial relationship is established between two components if the minimum Euclidean distance between their bounding boxes is within a predefined threshold (e.g., 0.3m). The specific choice of this threshold is discussed in Section 4.2.1.

In Table 3, the positioning coordinates refer to the coordinates of component positioning points or the endpoints of positioning lines. For instance, columns and fittings are positioned using point coordinates, while beams and ducts use centerline positioning. The offset represents the vertical deviation of a component's top or bottom relative to its elevation constraints (e.g., top or bottom elevations). For instance, the bottom of a column may be offset by 50 mm below its bottom elevation and 100 mm below its top elevation. A local spatial relationship identifies all components within a specified distance from a given component and calculates their positional relationships, including 1) spatial relationship, 2) angle, 3) shortest distance vector coordinates, 4) shortest distance, and 5) angle with the horizontal plane, as shown in Figure 3. The specific calculation methods are detailed in previous studies (Han et al., 2025). Introducing local spatial relationships provides additional geometric information and contextual context, enabling DL models to better understand the relative spatial arrangement of components and improving their ability to handle complex positional relationships with greater accuracy. Furthermore, since the range of local spatial relationships directly influences the number of graph nodes and the input data volume, it will be treated as a hyperparameter and analyzed in subsequent experiments to optimize performance.

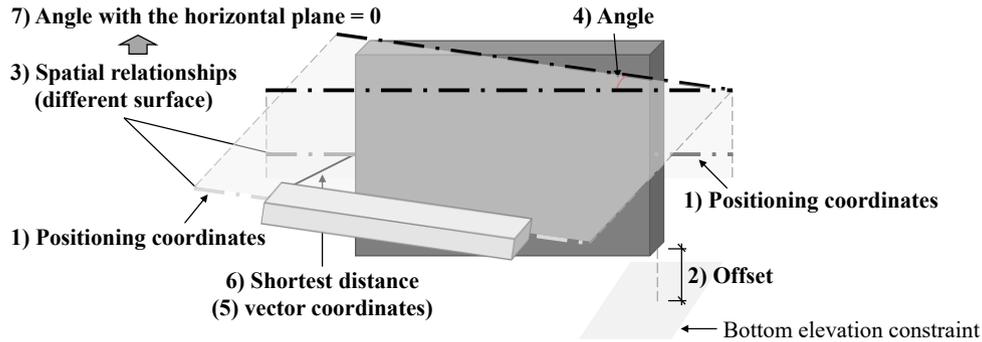

**FIGURE 3** Example of local spatial relationship.

### 3.1.4 | Feature processing and embedding for graph representations of BIM models

This section further embeds, aligns, and normalizes the components, their properties, and relationships extracted in Sections 3.1.1, 3.1.2, and 3.1.3 into graph nodes, node features, and edges, thereby achieving a BIM-specific scalable graph representation. It is important to note that graphs can be classified into homogeneous and heterogeneous graphs. In a homogeneous graph, all nodes represent the same type of entity, and all edges represent the same type of relationship. In contrast, a heterogeneous graph allows for different types of nodes and edges. The structure of the graph influences the message-passing mechanism of the neural network, which in turn significantly influences the model's learning performance. Therefore, this study will explore the impact of these different graph structures on model performance in subsequent experiments.

**(1) Construction method for the BIM-specific scalable graph representation**

First, the method for constructing heterogeneous graphs is introduced, with the construction process detailed in Table 4. This study represents each floor of different BIM models as a distinct graph. In contrast to the node and edge construction methods in previous studies (Han et al., 2025), this research introduces improvements, as shown in Figure 4. Figure 4(a) illustrates the network-based representation of BIM models as proposed by Han et al. (2025). The red annotations in this panel highlight the key improvements introduced in this study, which enable the transformation of the original network into a unified graph structure (Figure 4(b)). Specifically, to avoid excessive computational burden from too many node types, this study treats all component nodes in the original network as semantic nodes and consolidates three relationship types—Host, Connection, and Touch floor—into a unified topological relationship. Additionally, to convert the model checking task into a node classification task (detailed in Section 3.3), the edges representing spatial and topological relationships in the original network were treated as spatial and topological nodes, respectively. For example, if there is a topological or spatial relationship between two components, the semantic nodes representing these components are connected to the corresponding relationship nodes via edges. Notably, since two components with a topological relationship are adjacent, with a relative distance of zero, a spatial relationship inherently exists. When the topological relationship is incorrect, both relational nodes would need to be classified as errors, as this also implies an error in the distance between components and, consequently, an incorrect spatial node feature. This redundancy increases the difficulty of node classification tasks. Therefore, the spatial relationship is ignored when a topological relationship exists, ensuring that the two component nodes are connected by only one relational node. Furthermore, as the semantic, spatial, and topological features of all BIM components are represented by graph nodes, the edges in the graph do not contain specific features and simply denote basic information transfer relationships.

**TABLE 4** Process of constructing the BIM-specific graph.

| Algorithm |
|---|
| Input: Network $\mathcal{N}\{\mathcal{V}, \mathcal{E}\}$ |
| Output: Graph $\mathcal{G}\{\mathcal{V}, \mathcal{E}\}$; node features $\{\mathbf{n}_v, \forall v \in \mathcal{V}\}$ |
| 1:    For each component node $i$ in $\mathcal{N}$ do |
| 2:      $v_i^{Se}$ = create_node($i$) |
| 3:      $\mathcal{F}_{v_i^{Se}}$ = filter_semantic_features($i$) |
| 4:      $\mathbf{n}_{v_i^{Se}}$ = encode_and_alignment_features($\mathcal{F}_{v_i^{Se}}$) |
| 5:    end for |
| 6:    For each (Host, Connection, and Touch floor) edge $e$ in $\mathcal{N}$ do |
| 7:      component node i, j = extract_endpoint_node($e$) |
| 8:      $v_{i,j}^T$ = create_node(i, j) |
| 9:      $(v_i^{Se}, v_{i,j}^T)$ = create_edge($v_i^{Se}, v_{i,j}^T$) |
| 10:    $(v_{i,j}^T, v_j^{Se})$ = create_edge($v_{i,j}^T, v_j^{Se}$) |
| 11:    $\mathbf{n}_{v_{i,j}^T}$ = topological_relationship_type($e$) |
| 12:    end for |
| 13:    For each spatial edge $e$ in $\mathcal{N}$ do |
| 14:    i, j = extract_endpoint_node($e$) |
| 15:    if not $v_{i,j}^T$: |
| 16:      $v_{i,k}^{Sp}$ = create_node(i, k) |

```
17:     (v_i^Se, v_{i,k}^Sp) = create_edge(v_i^Se, v_{i,k}^Sp)
18:     (v_{i,k}^Sp, v_k^Se) = create_edge(v_{i,k}^Sp, v_k^Se)
19:     n_{v_{i,k}^Sp} = extract_features(e)
20:   end for
      V ← ∀{v_i^Se, v_{i,j}^T, v_{i,k}^Sp}
21:   E ← ∀{(v_i^Se, v_{i,j}^T), (v_{i,j}^T, v_j^Se), (v_i^Se, v_{i,k}^Sp), (v_{i,k}^Sp, v_k^Se)}
      n_v ← ∀{n_{v_i^Se}, n_{v_{i,j}^T}, n_{v_{i,k}^Sp}}
```

$$22: \quad \mathbf{n}_v \leftarrow \frac{\mathbf{n}_v}{max(|max(\mathbf{n}_v)|, |min(\mathbf{n}_v)|)}$$

Where $v_i^{Se}$, $v_{i,j}^T$, and $v_{i,k}^{Sp}$ denote semantic, topological, and spatial nodes, respectively. It should be noted that in homogeneous graphs, these node types are not distinguished; the distinction in the table is provided solely for clarity. $\mathcal{V}$ represents all nodes and $\mathcal{E}$ represents all edges in the graph.

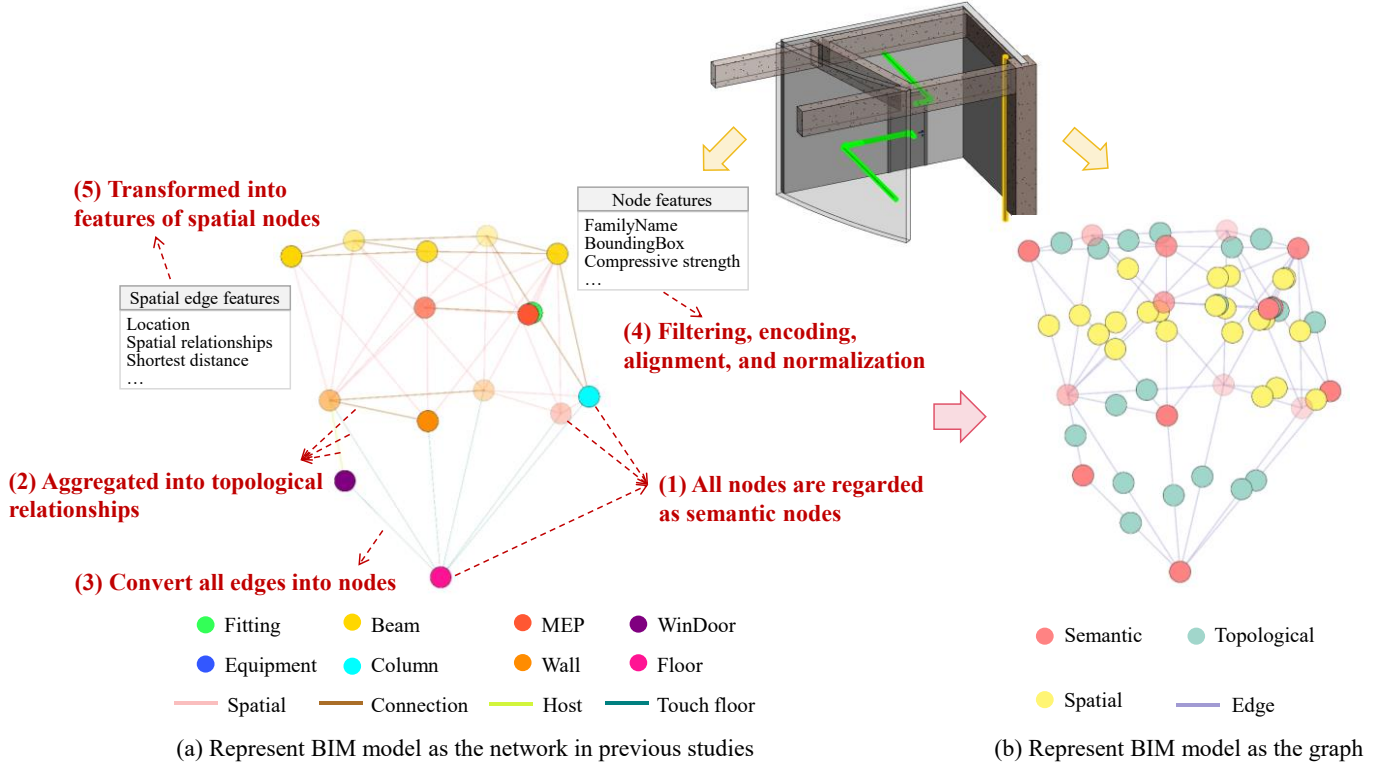

**FIGURE 4**    The difference between converting the BIM model into a network and a graph. Node positions in the figure are determined by the coordinates of their corresponding components' center points.

### (2) Embedding the semantic-spatial-topological features of BIM components into graphs

Finally, the semantic, topological, and spatial properties of components extracted in Sections 3.1.1, 3.1.2, and 3.1.3 were converted into features for different types of nodes, as illustrated in Table 4. For the topological properties, they are represented by a one-hot encoding of their respective topological type. For the spatial properties, since these values are numerical, no embedding is required. For the semantic properties, further embedding is performed to integrate them into the semantic node features. Specifically, 1) shape and structural purpose features are embedded using a one-hot method; 2) dimensions features are expressed as numerical values in millimeters, requiring no additional embedding; and 3) family and family type features, which are typically discrete and variable-length text data, are embedded into fixed-dimensional vector representations using the Moka Massive Mixed Embedding (M3E). This method can effectively capture the semantic meaning and relationships of these names, ensuring that similar components are represented with proximity in the vector space. However, if the embedding dimension of family and family type features is too high, it may hinder the model's ability to learn other property features effectively. Conversely, if the dimension is too low, it may fail to adequately capture the complexity and variability of the data, potentially leading to loss of critical information and reduced model performance. To determine the optimal configuration, model performance was evaluated with embedding dimensions of 32, 64, 256, 512, and 768. Ablation experiments showed that a 64-dimensional embedding achieved the best results, which was adopted for all subsequent experiments.

Given the potential absence of certain properties for some components, alignment is necessary. Alignment refers to cases where certain components may lack specific properties or their values may not be extractable, requiring placeholders of 0 to maintain consistency. For example, wall components have two valid coordinates corresponding to the endpoints of their positioning line, whereas column components only have a single valid coordinate for the positioning point. To ensure alignment with the wall, the column's positioning coordinates property is structured so that the first three entries represent the positioning point coordinates, while the remaining three entries are filled with 0 as placeholders. The lengths of each property after alignment are shown in Table 3. Additionally, due to the large differences in the magnitudes of various properties, normalization was applied to improve the model's training convergence speed and stability. Specifically, the normalization process was conducted per floor plan, scaling the relevant property values to the range of -1 to 1 while ensuring that 0 values remained unchanged, as described in Equation (1).

$$X'_i = \frac{X_i}{max(|max(X_i)|, |min(X_i)|)} \quad (1)$$

Where $X_i$ represents the i-th property value of all components on the same floor plan, and $X'_i$ denotes the normalized value.

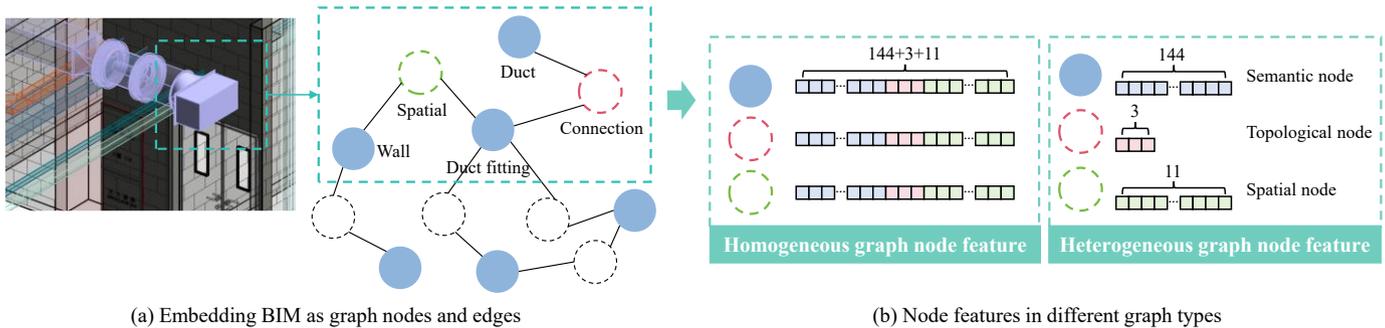

(a) Embedding BIM as graph nodes and edges     (b) Node features in different graph types

**FIGURE 5**   Graph representations of BIM models.

In heterogeneous graphs, three types of nodes are distinguished: semantic, topological, and spatial, each with distinct feature sets, as shown in Figure 5. The feature of each semantic node is a concatenated vector of five semantic properties and the first two spatial properties, resulting in a (1, 144)-dimensional vector. The features of topological nodes correspond to the topological properties, which, in this study, results in a (1, 3)-dimensional vector, as there are three types. The feature of spatial nodes is a concatenated (1, 11)-dimensional vector formed by embedding the remaining five spatial properties from Table 3. For homogeneous graphs, the only difference in graph construction compared to heterogeneous graphs is that all nodes are treated as the same type, with identical dimensional features, as illustrated in Figure 5. Specifically, all properties from Table 3 are embedded and concatenated into a (1, 158)-dimensional vector. If any node lacks certain properties, the missing values are replaced with zeros. For example, semantic nodes do not include topological or spatial properties, so the corresponding positions in the node feature vector are filled with 0.

### 3.1.5 | Dataset development for pre-training

To enable the pretrained large foundation model to learn design experiences from different building types and design firms while ensuring dataset diversity and representativeness, this study selected BIM models from multiple building types, including hospitals, residential, and industrial buildings. Since some models resulted in an excessively large number of nodes after conversion into graph form, directly feeding them into a GNN could lead to challenges such as increased computational complexity and memory requirements. Therefore, some models were divided into several non-overlapping regions to reduce the graph size while preserving the spatial relationships between components in the local areas. The BIM-specific scalable graph representation method, as described in Section 3.1, was then applied to convert the collected models into graphs, resulting in a BIM-specific graph dataset comprising 176 graphs, with a total of 974,991 nodes and 3,414,280 edges. The distribution of the dataset is shown in Figure 6.

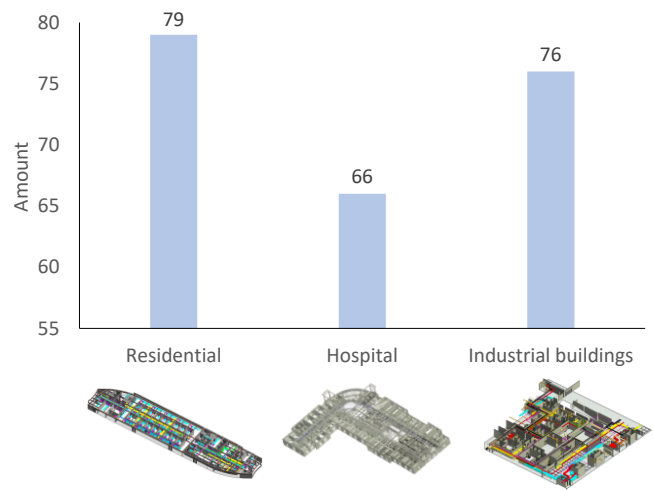

**FIGURE 6**   Characteristics of pre-training datasets.

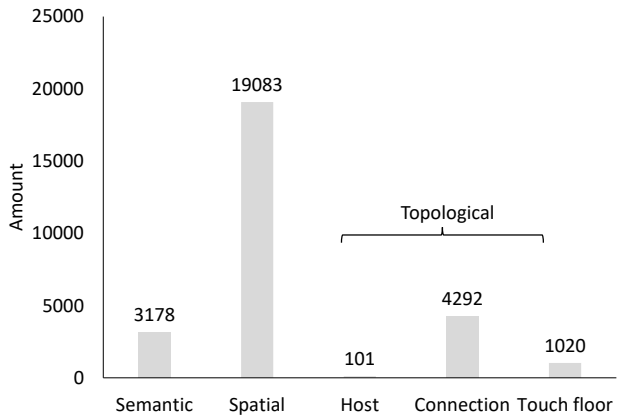

**FIGURE 7** Node type counts of a floor of a building represented as a graph.

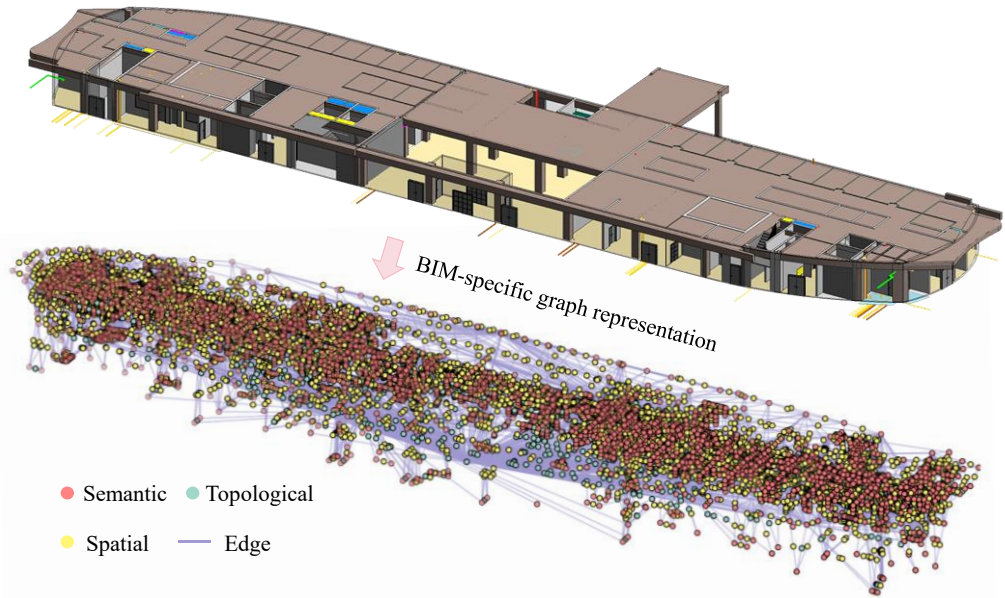

**FIGURE 8** An example of a segment of a building's BIM model represented as a graph.

An example of representing a floor of a BIM model as a heterogeneous graph (with a local spatial relationship range of 0.3 m) is shown in Figure 8, with the node count for each type depicted in Figure 7. The conversion of BIM models into graph representations was performed on a workstation equipped with an Intel Core i9-13900H CPU, 32 GB RAM, 1 TB SSD, and an NVIDIA RTX 4060 GPU. The process was implemented using Python 3.9 and Revit API. For BIM models created in Revit, the conversion to a graph structure typically requires approximately five minutes per floor, with actual times varying based on model complexity and the number of extracted relationships. It is important to note that, for a homogeneous graph, node types will not be distinguished.

### 3.2 | Architecture of the proposed BIGNet

This study builds upon and extends the GraphMAE2 architecture (Hou et al., 2023) for pretraining the first large-scale GNNs to learn implicit design features of components. The architecture was improved to broaden its applicability, making it suitable for both homogeneous and heterogeneous graphs. By masking certain nodes in the input graph, the pretraining task is transformed into a node feature decoding and reconstruction problem. This approach enables the model to recover the entire graph structure from local information through self-supervised learning, as illustrated in Figure 9. Notably, our pretrained model is task-agnostic and can be efficiently adapted to multiple distinct BIM downstream tasks within a unified framework. And the term "large-scale" does not refer to the size of a single graph (i.e., the number of nodes or edges in one BIM project's representation), but rather to the overall scale of the dataset, which encompasses a broad collection of BIM models drawn from diverse projects, building types, and design contexts.

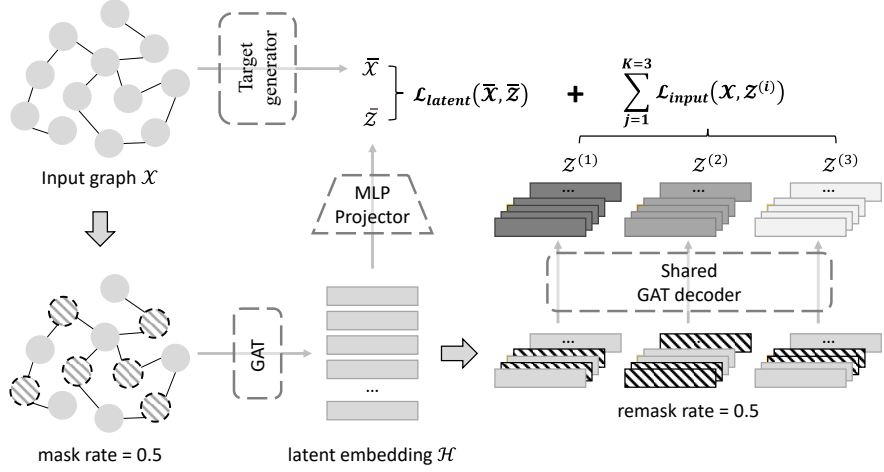

**FIGURE 9** Overview of GraphMAE2 framework.

### 3.2.1 | Masking strategy

Specifically, the architecture consists of an encoder and a decoder. The pretraining goal is to obtain a well-initialized encoder by reconstructing the input node features. The main task of the encoder is to reconstruct the graph with partially masked nodes by capturing the local structure (i.e., the features of the unmasked nodes and their neighborhoods), thus generating a high-dimensional representation $\mathcal{H}$ that captures global features. The decoder then uses this representation $\mathcal{H}$ to reconstruct the features of the masked nodes, based on two decoding strategies: 1) Multi-view random re-mask decoding to mitigate overfitting to input features, and 2) Latent representation prediction to capture more information.

The multi-view random re-mask decoding refers to the process of randomly re-masking the embedded high-dimensional features $\mathcal{H}$ during decoding. Specifically, different nodes are randomly masked multiple times, and a shared decoder reconstructs the input features. After hyperparameter tuning, this study uses a relatively high masking rate of 50% during both encoding and decoding. The features of masked nodes are randomly filled, and three random re-mask views are generated for decoding. Finally, scaled cosine error (Hou et al., 2022) was used to measure reconstruction error, summing the errors from the three views for training.

$$\mathcal{L}_{input} = \frac{1}{|\mathcal{V}|} \sum_{j=1}^{K=3} \sum_{v_i \in \mathcal{V}} \left(1 - \cos(\theta_{ij})\right) \quad (2)$$

$$cos(\theta_{ij}) = \frac{x_i^T z_i^{(j)}}{\|x_i\| \cdot \left\|z_i^{(j)}\right\|} \quad (3)$$

Where V represents the input masked nodes, and K denotes the number of random re-mask views, set to 3 in this study. $x_i$ refers to the $i$-th row of the input features, $z_i^{(j)}$ represents the $i$-th row of the predicted feature $Z^{(j)}$ under the $j$-th re-mask view, and $\cos(\theta_{ij})$ denotes the cosine of the angle between the vectors $x_i$ and $z_i^{(j)}$.

Latent representation prediction aims to design an additional prediction task while minimizing the direct influence of input features. Therefore, prediction is performed in the representation space rather than the input feature space. This part involves three networks: the target generator, encoder, and projector. The target generator generates a latent prediction target $\bar{X}$ from the unmasked graph, sharing the same architecture as the encoder and projector but with different weights. Meanwhile, the encoder's output $\mathcal{H}$ is projected into the representation space by the projector, yielding the latent prediction $\bar{Z}$. The encoder is trained by minimizing the distance between $\bar{Z}$ and $\bar{X}$.

$$\mathcal{L}_{latent} = \frac{1}{N} \sum_{i}^{N} \left(1 - \cos(\bar{\theta}_{ij})\right) \quad (4)$$

$$\cos(\bar{\theta}_{ij}) = \frac{\bar{z}_i^T \bar{x}_i}{\|\bar{z}\| \cdot \|\bar{x}\|} \quad (5)$$

$$\zeta \leftarrow \tau\zeta + (1-\tau)\xi \quad (6)$$

Where N represents the number of nodes, $\bar{x}_i$ is the $i$-th row of $\bar{X}$, $\bar{z}_i$ is the $i$-th row of $\bar{Z}$, $\cos(\bar{\theta}_{ij})$ denotes the angle between two vectors. The learnable weights $\zeta$ of the target generator are updated via an exponential moving average of the learnable weights $\xi$ from the encoder and projector, with weight decay $\tau$.

The final loss function is obtained by summing the two losses from the training process:

$$\mathcal{L} = \mathcal{L}_{input} + \mathcal{L}_{latent} \quad (7)$$

In this study, since there is always a topological or spatial node between two semantic nodes with a specific relationship, both the encoder and decoder are implemented as 2-layer GAT (Graph Attention Network) networks. This ensures that the features of surrounding component nodes are considered during both encoding and decoding while minimizing computational costs and facilitating effective information flow.

### 3.2.2 | Message-passing mechanism for different graph structures

This study further explores how model performance is affected by representing input graphs as either homogeneous or heterogeneous structures. While the original GraphMAE2 model by Hou et al. (2023) was developed exclusively for homogeneous graphs, the current work extends its application to heterogeneous graphs by introducing targeted improvements to both the masking strategy and the message-passing mechanism. Specifically, distinct masking rates are assigned to different node types: a higher masking rate (0.6) is applied to spatial and topological nodes, which typically involve complex, multi-dimensional information such as coordinates, distances, and shapes, to strengthen the model's capacity for learning deep features. For semantic nodes, the masking rate remains at 0.5, consistent with the homogeneous setting. The effect of varying masking rates by node type warrants further exploration in future research.

Additionally, the message-passing mechanism has been refined to accommodate the differing informational distributions and representational needs of each node and meta-path type. This is achieved by assigning dedicated attention heads and weights to each category. To better reflect the physical relationships between components, the revised message-passing strategy is illustrated in Figure 10.

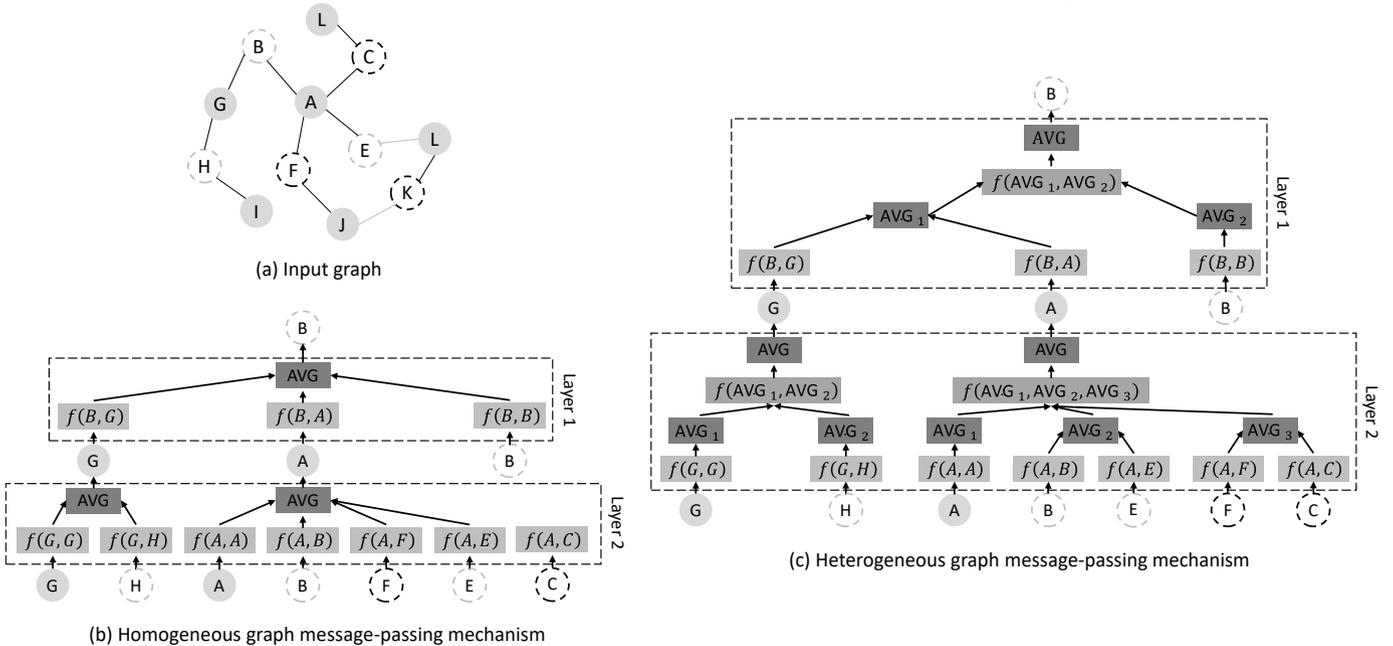

**FIGURE 10**    Two-layer GAT example with message-passing mechanism.

The first pretrained graph neural network, BIGNet, obtained by pre-training using the above node reconstruction method, can effectively guide the encoder to focus on the topological and spatial features of neighboring nodes. This enables the model to capture the implicit design features within BIM data more deeply. Additionally, it strengthens the encoder's global feature representation, providing a more robust foundation for subsequent node classification tasks and facilitating transfer learning in small-sample data tasks. Notably, during downstream tasks, the decoder and projector are discarded, with only the encoder used to generate embeddings or fine-tune.

### 3.3 | Evaluation of BIGNet in different transfer learning tasks

To investigate the impact of different graph representations and transfer learning methods on the performance of BIGNet, and to assess whether the proposed BIGNet effectively learns the underlying universal component characteristics and intrinsic relationships between components in design, this study designs multiple BIM-based design checking tasks based on transfer learning, leveraging the knowledge acquired during pretraining.

### 3.3.1 | Transfer learning dataset construction

To assess the effectiveness and generalizability of BIGNet, this study introduces three common types of errors frequently encountered in real-world projects: 1) Semantic conflicts: The semantic features of a component do not meet standard design requirements. For example, some walls are incorrectly created using beam families, which will lead to cascading effects in subsequent processes such as cost estimation and operation/maintenance, increasing both cost deviations and maintenance complexity. 2) Data range error: The values of a component's semantic features fall outside regulatory standards. For instance, some door heights do not meet the required range, with standard industry guidelines specifying heights between 200 cm and 240 cm. 3) Topological error: Errors or omissions in the topological relationships between components. This can occur when MEP components lose connections due to changes in distance or attributes, often resulting from modifications to the layout. It is important to note that the graph representation methods proposed in this study allow the conversion of error detection in BIM models into a node classification problem, as shown in Figure 11. The first two error types, related to the component properties, are associated with semantic nodes. The third type concerns the relationship between two component nodes, specifically spatial or topological nodes. For example, a connection node exists between an MEP component and its corresponding fitting component. If this connection is lost, a spatial node will replace it. Such errors typically occur when the position or connection relationship is not restored during a drawing revision, leading to slight positional discrepancies but remaining within the local spatial relationship range. Therefore, when representing the BIM model as a graph, the node that should have been a connection edge will be replaced by a spatial node.

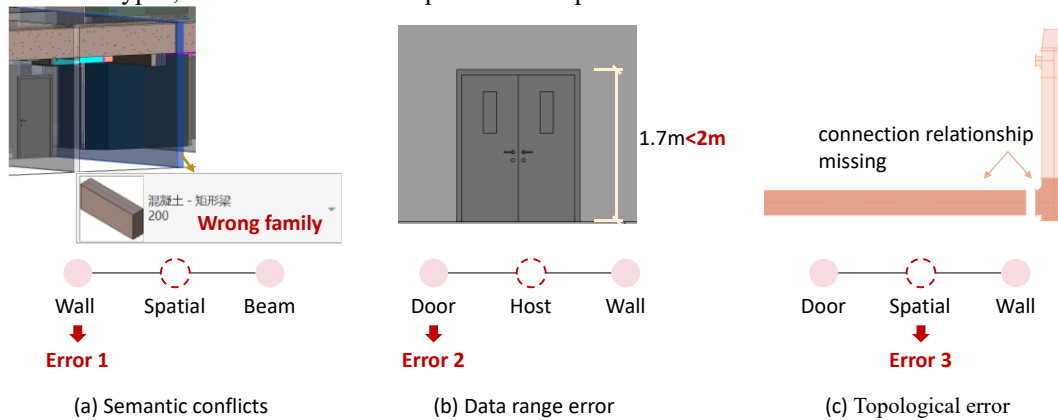

FIGURE 11  Convert BIM inspection issues into node classification tasks.

To validate the feasibility and generalization of the proposed method, this study annotated 16 BIM models from real-world projects that span a range of architectural functions, including hospitals, residential complexes, and industrial facilities. Each BIM model corresponds directly to the final design drawings used for the actual buildings and incorporates comprehensive information covering architectural, structural, and MEP systems. All models were developed using Autodesk Revit and adhere to industry standards, containing accurate geometric representations, material specifications, family classifications, and detailed semantic attributes. This ensures that each model offers a high level of detail, making them well-suited for advanced analysis, such as design checking and error annotation.

Specifically, 30% of the wall types, door heights, and the relationships between MEP components and their fittings in the model were modified and labeled as errors. To adapt to the error detection tasks, the dataset was divided according to different proportions, as described in Section 4.1. When the training set for transfer learning comprised 30% of the total transfer dataset, five regions were used for transfer learning, five for validation, and six for testing. The distribution of error categories across the training, validation, and test sets is illustrated in Figure 12. Notably, the imbalance in datasets (as shown in Figure 12), where topological errors outnumber the other two categories, primarily arises from the inherent composition of BIM models and the data construction process. In our three transfer learning tasks, the number of wall and door components involved in Semantic conflicts and data range errors is substantially lower than the number of connection relationships between MEP components and their fittings relevant to topological errors. To better simulate real-world conditions, we modified and labeled the same proportion of each relevant component or relationship type in the model as errors. As a result, the actual number of topological errors became significantly higher than that of the other two error types. Additionally, topological errors are intrinsically more challenging, as their detection requires the model to synthesize spatial coordinates, dimensions, and contextual information to infer the validity of inter-component connections. Therefore, the larger proportion of such errors in the dataset is reasonable and beneficial, as it provides the model with greater opportunities to learn and generalize from these complex relationships.

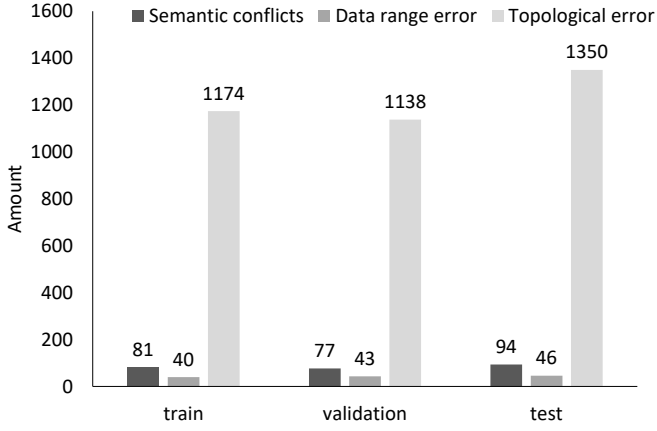

**FIGURE 12** The number of each error type in the test projects.

### 3.3.2 | Different graph representation methods and transfer learning strategies

As discussed in Section 3.1, the main factors influencing the BIGNet are: 1) the range of local spatial relationships (tested at 0.2, 0.3, 0.4, 0.5, and 0.6), and 2) the type of graph structure (homogeneous vs. heterogeneous). Therefore, this study further investigates the impact of these parameters in the experiments to identify the optimal graph representation methods.

The transfer learning methods for BIGNet include feature extraction and fine-tuning. In feature extraction, the encoder of BIGNet is used as a feature extractor without modifying its parameters. The high-level features extracted are directly applied to the new task, followed by training a new classification model on these features to make predictions. In fine-tuning, both the encoder of BIGNet and its parameters are updated through further training to better adapt to the new task. In transfer learning tasks, MLPs are commonly used as classifiers due to their computational efficiency and ease of implementation and optimization. However, if structural information in the data is important, MLPs may not capture it effectively. Therefore, this study also compares the performance of a GAT classifier. In summary, the study investigates the performance of three transfer learning methods: 1) Feature extraction followed by an MLP classifier, as shown in Figure 13(a); 2) Feature extraction followed by a GAT classifier, as shown in Figure 13(b); and 3) Fine-tuning followed by an MLP classifier, as shown in Figure 13(c).

It is important to note that, for the three error types considered in this study, the transfer-learned model will classify nodes into four categories: correct, semantic conflicts, data range error, and topological error. Given the significant class imbalance in the dataset, a weighted cross-entropy loss function is employed, with adaptive adjustments based on the model's prediction error rates. If the misclassification rate for a specific class is high, its weight is increased to encourage the model to focus more on that class during each training cycle.

$$w_i(t+1) = w_i(t) \times (1 + \alpha \times \text{Error}_i) \quad (8)$$

Where $\alpha$ is a tuning factor, set to 0.1 in this study, and $\text{Error}_i$ represents the misclassification rate of class $i$ during the current training cycle.

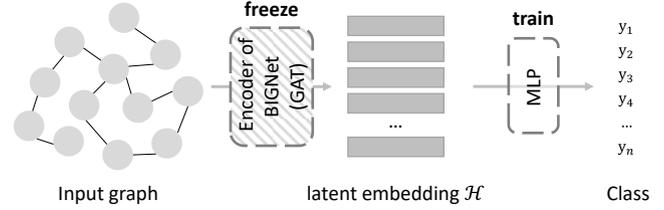

(a) Feature extraction followed by an MLP classifier

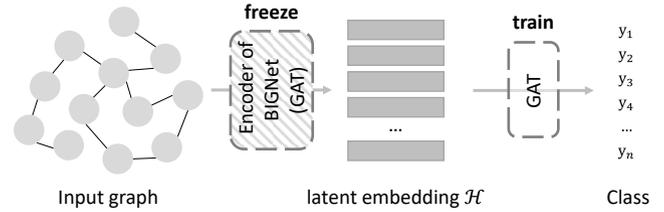

(b) Feature extraction followed by a GAT classifier

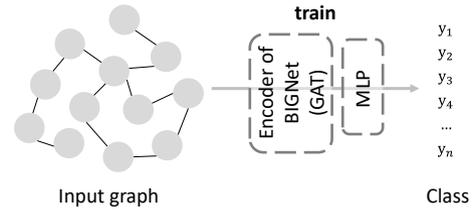

(c) Fine-tuning followed by an MLP classifier

**FIGURE 13** Three transfer learning strategies.

### 3.3.3 | Evaluation metrics

To assess the effectiveness of the proposed method for BIM-based design checking tasks, a typical classification problem, this study uses the F1-score as the evaluation metric, derived from the confusion matrix as well as precision and recall. The confusion matrix is a square matrix where rows represent predicted classes and columns represent actual classes. Precision measures the proportion of true positives among predicted positives, while recall indicates the proportion of true positives correctly identified. The F1-score, the harmonic mean of precision and recall, provides a balanced evaluation of model performance.

$$\text{Precision} = TP / (TP + FP) \quad (9)$$
$$\text{Recall} = TP / (TP + FN) \quad (10)$$
$$\text{F1 score} = 2 \times (\text{precision} \times \text{recall}) / (\text{precision} + \text{recall}) \quad (11)$$

Where TP denotes the number of true positive samples predicted as positive, FP represents the number of true negative samples predicted as positive, and FN refers to the number of true positive samples predicted as negative.

It is important to note that due to the large number of correct nodes in the drawings, and the fact that most models can accurately predict the majority of correct nodes, the weighted F1-scores are generally above 0.97, with minimal numerical differences, which fails to adequately highlight the performance variations between models. Furthermore, since the downstream task of this study is model checking, the key measure of performance is the model's ability to accurately classify erroneous nodes. Therefore, the Average F1-score is used as the metric to evaluate model performance on this task.

## 4 | EXPERIMENTS AND RESULTS

### 4.1 | Experiment settings

Based on Section 3.3, this study examines two key influencing factors: 1) Graph representation methods and transfer learning training set size, with specific experimental parameters detailed in Table 5 and Table 6, and 2) Transfer learning methods, with parameters outlined in Table 7. Additionally, since graph representation and pre-training are independent processes, this study does not consider their potential coupling effects. Finally, to validate the effectiveness of the proposed BIGNet, a baseline without pre-training is also included as a control group.

**TABLE 5** Experiments on graph representation methods.

| Value range of the 'spatial' edge (/m) | Homogeneous graph | Heterogeneous graph |
|---|---|---|
| 0.2 | Ho_S0.2 | He_S0.2 |
| 0.3 | Ho_S0.3 | He_S0.3 |
| 0.4 | Ho_S0.4 | He_S0.4 |
| 0.5 | Ho_S0.5 | He_S0.5 |
| 0.6 | Ho_S0.6 | He_S0.6 |

**TABLE 6** Experiments varied the proportion of the dataset used for transfer learning.

| Transfer learning training set ratio | Homogeneous graph | Heterogeneous graph |
|---|---|---|
| 0.1 | Ho_0.1 | He_0.1 |
| 0.2 | Ho_0.2 | He_0.2 |
| 0.3 | Ho_0.3 | He_0.3 |
| 0.4 | Ho_0.4 | He_0.4 |
| 0.5 | Ho_0.5 | He_0.5 |

**TABLE 7** Experiments on transfer learning methods.

| Transfer learning methods | Homogeneous graph | Heterogeneous graph |
|---|---|---|
| Feature extraction with MLP | Ho_BIGNet_MLP | He_BIGNet_MLP |
| Feature extraction with GAT | Ho_BIGNet_GAT | He_BIGNet_GAT |
| Fine-tuning | Ho_BIGNet_Tune | He_BIGNet_Tune |
| No pretrain with MLP | Ho_NoPre_MLP | He_NoPre_MLP |
| No pretrain with GAT | Ho_NoPre_GAT | He_NoPre_GAT |

For the GAT models used as the encoder and decoder, hyperparameters were selected based on prior research (Hou et al., 2023). The number of hidden layers was set to 2, with a hidden layer dimension of 512, 4 attention heads in the intermediate layers, and 1 attention head in the output layer. Furthermore, the impact of different initial learning rates (0.001, 0.003, 0.005, and 0.007) and batch sizes (4, 8, and 16) was considered by grid search, and a cosine decay strategy was applied for learning rate scheduling. The optimizer is Adam, and the max epoch was set to 5000. Besides, to reduce training time, an early stop mechanism was added. If the model performance on the checking dataset does not improve for 300 epochs, the training process will be terminated.

The computing platform used for the experiments is configured as follows: Windows Server 2019 Standard as the operating system, an Intel Xeon E5-2682 v4 CPU running at 64 cores with a clock speed of 2.5 GHz, 54 GB of RAM, and an NVIDIA GeForce RTX 3090 GPU with 24 GB of memory.

### 4.2 | Experiment results

#### 4.2.1 | Performance comparison of different graph representation methods

**(1) The impact of different graph structures on model performance**

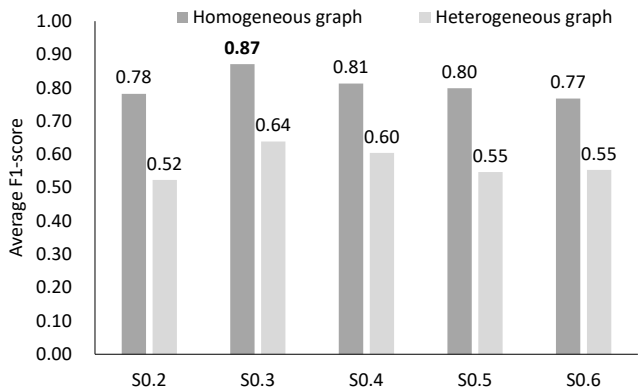

**FIGURE 14** Average F1-score on different graph representation methods with the transfer learning training set accounting for 30% of the transfer dataset.

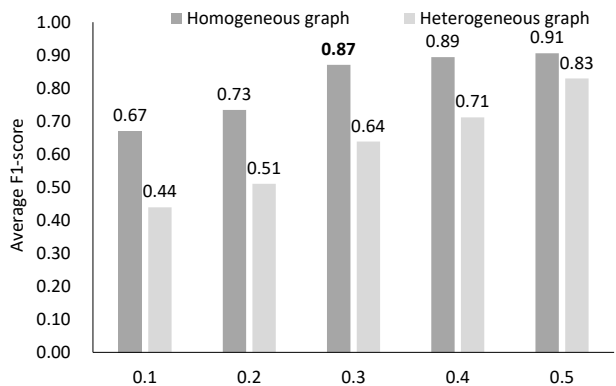

**FIGURE 15** Average F1-scores for models under varying proportions of training data.

Figure 14 presents the Average F1-scores for models using different graph representation methods on the three BIM-based design checking tasks, with the transfer learning training set accounting for 30% of the transfer dataset as described in Section 3.3.1. Figure 15 shows the Average F1-scores for models under varying proportions of training data.

The results indicate that homogeneous graphs achieve higher accuracy than heterogeneous graphs under the same training data conditions in this study. This outcome can be attributed to several theoretical factors. First, homogeneous graph representations treat all nodes as the same type, resulting in a substantially reduced parameter space compared to heterogeneous graphs, which require separate parameters (such as attention heads and weights) for each node and edge type. While heterogeneous graphs can capture richer and more diverse information by distinguishing between node types, this added complexity increases both the number of parameters and the difficulty of training. Given that the objective of this research is to transfer pretrained models to specific tasks using limited training data, the simpler structure and fewer parameters of homogeneous graphs offer a significant advantage, enabling more effective training under data-scarce conditions.

Although homogeneous graphs achieve marginally better performance when the training set proportion increases from 0.3 to 0.4 or 0.5, the improvement is not sufficient to justify the additional labeling workload. Therefore, a 30% training set proportion is considered optimal for this study. Notably, the consistent increase in F1-score for heterogeneous graphs as the training set size grows further suggests that, with sufficient data in future applications, heterogeneous graphs may ultimately provide greater expressive power and superior downstream performance. In addition, training heterogeneous graphs incurs substantially higher computational costs—approximately five times longer than homogeneous graphs—further supporting the efficiency and suitability of homogeneous graphs when data and computational resources are limited.

**(2) The impact of the value range of spatial relationships on model performance**

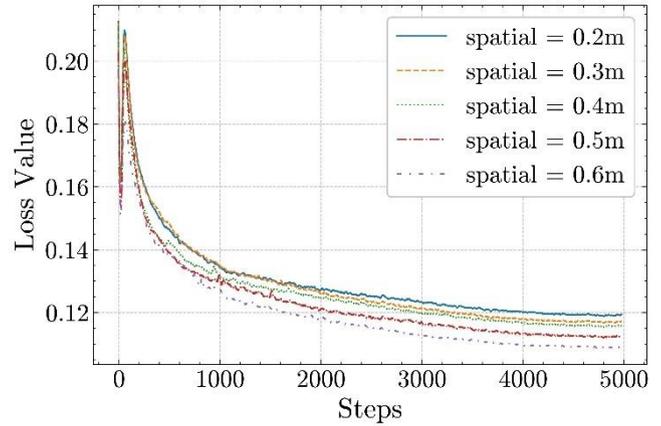

**FIGURE 16** Pre-training loss curve of the homogeneous graph.

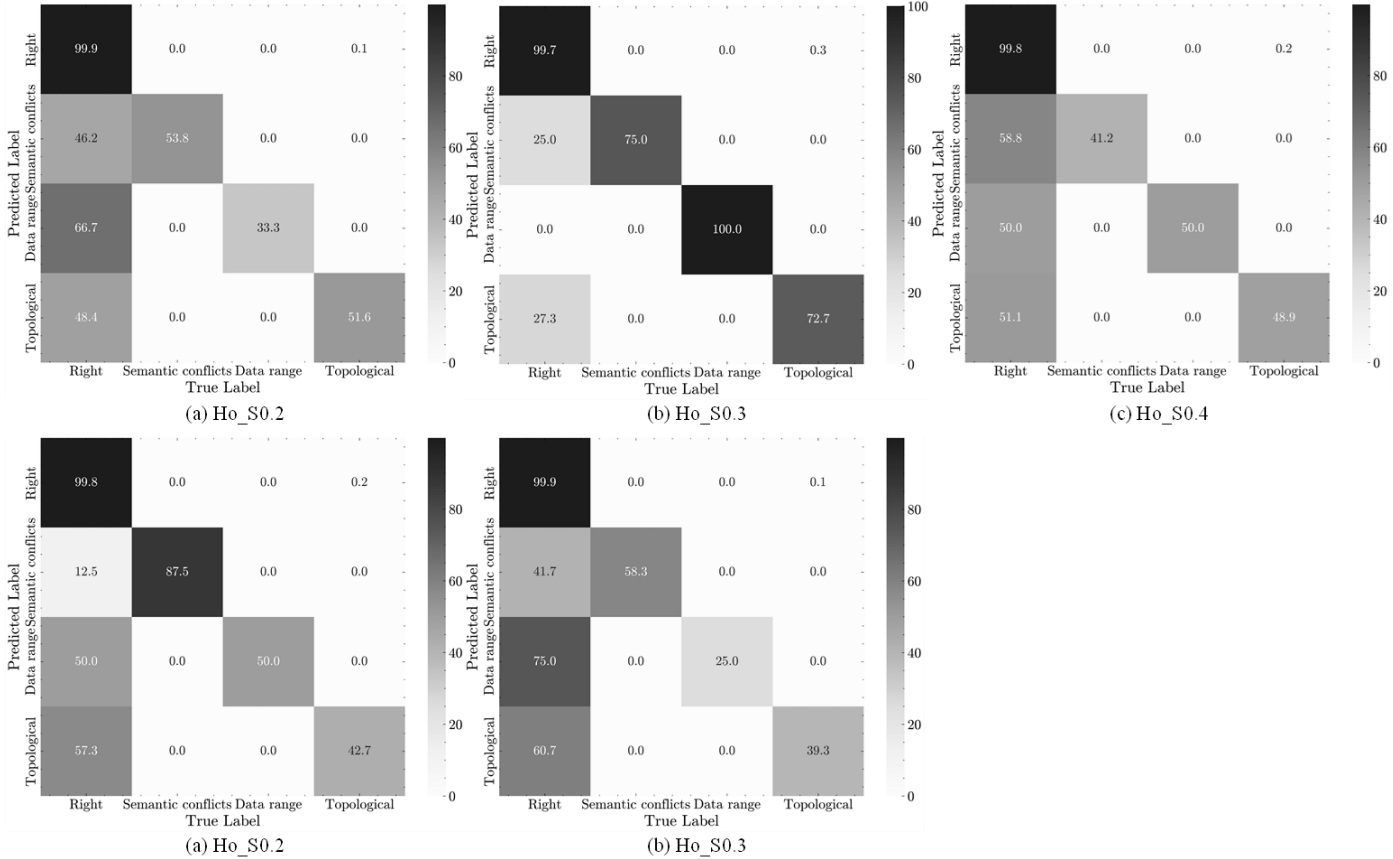

FIGURE 17 Confusion matrix for a specific test graph under different 'spatial relationship' value ranges.

When representing BIM models as homogeneous graphs using different local spatial relationship ranges, the pretraining loss curves are illustrated in Figure 16. To highlight distinctions between the curves, the plot starts at step 10. The results indicate that the final loss value decreases progressively as the local spatial relationship range increases. This trend may be attributed to the larger local spatial relationship range providing richer relative positional information about neighboring nodes, which helps BIGNet capture more intricate semantic and spatial features between nodes. Consequently, the model becomes more effective at reconstructing the masked node features.

However, as shown in Figure 14, Ho_S0.3, rather than Ho_S0.6, achieved the best performance in design checking. To further clarify the differences in prediction outcomes, a confusion matrix for a sample from the test set is shown in Figure 17. This result may stem from the trade-off introduced by the increased local spatial relationship range: while a larger range provides more positional information, it also increases the number of nodes in the graph, exacerbating the imbalance between the number of correct and erroneous nodes and increasing the difficulty of transfer learning. Conversely, if the local spatial relationship range is too small, the model lacks sufficient contextual information to make accurate predictions. Thus, selecting an appropriate local spatial relationship range (0.3m in this study) can effectively balance information richness and complexity, resulting in higher accuracy. Accordingly, subsequent experiments will fix the local spatial relationship range at 0.3m.

### 4.2.2 | Performance comparison of different transfer learning methods

The Average F1-scores for model checking tasks, achieved using different transfer learning strategies after pretraining with homogeneous or heterogeneous graphs, are shown in Figure 18. The results show that fine-tuning achieves the highest performance on heterogeneous graphs. This is likely because the limited amount of transfer learning data is insufficient to train a classification model from scratch, making fine-tuning on pretrained parameters more suitable for heterogeneous graphs. Moreover, models based on heterogeneous graphs consistently underperform compared to the best-performing models using homogeneous graphs, which aligns with the analysis in Section 4.2.1. Consequently, the subsequent analysis will primarily focus on the results from training with homogeneous graphs.

To better understand the performance differences among various transfer learning methods, the confusion matrix of a sample from the test set was analyzed, as shown in Figure 19. The results show that Ho_BIGNet_GAT achieved the best performance, which utilizes the encoder from BIGNet as a feature extractor while training a new GAT model for classification. Its superior performance can be attributed to the significant differences between the pretraining and transfer learning tasks; directly fine-tuning BIGNet's parameters might lead to incorrect guidance. Conversely, as discussed in Section 3.2, pretraining enables BIGNet's encoder to effectively aggregate the topological structure and spatial features of neighboring nodes. Consequently, leveraging it as a feature extractor maps node features into a high-dimensional space can enhance global feature representation and provide a more robust foundation for subsequent node classification tasks. Additionally, the relatively smaller parameter size of the classification model designed for homogeneous graphs makes it better suited for transfer learning scenarios with limited data. Furthermore, Figures 18 and 19 also reveal that GAT consistently outperforms MLP. This highlights that the task's heavy reliance on capturing extensive topological and spatial relationships makes MLP inadequate for effectively learning these interactions. Therefore, for future applications, employing the Feature Extraction with GAT approach is recommended to transfer the design knowledge gained by BIGNet to specific tasks.

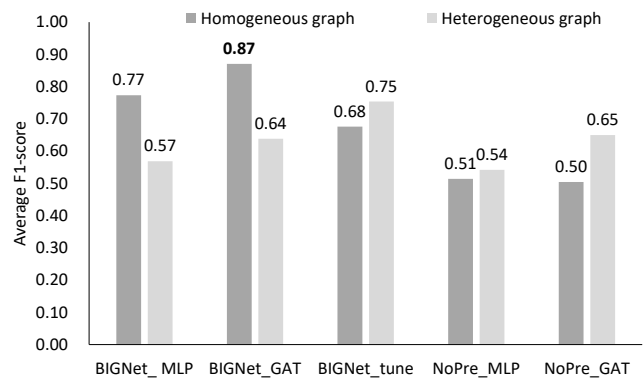

FIGURE 18 Experiment results on different transfer learning methods.

A detailed analysis of the upper and lower triangular sections of the confusion matrix in Figure 19 reveals that, in most cases, the proposed method misclassifies correct nodes as erroneous rather than misclassifying erroneous nodes as correct. This indicates that the model is more stringent in identifying erroneous samples, enhancing its reliability in practical applications, where false positives (erroneously predicting a correct node as accurate) could lead to more severe consequences.

In addition, the proposed method was compared with non-pretrained approaches to demonstrate the advantages of the pretraining and transfer learning framework. As shown in Figure 18, the proposed method achieves an Average F1-score improvement of 59.5% (with an MLP classifier) and 72.7% (with a GAT classifier) over traditional non-pretrained methods in BIM-based design checking tasks.

Figure 19 further reveals that while all models perform well on the first two simpler error types, their accuracy declines for the third, more complex error type. This is because the first two tasks primarily involve identifying deviations in component attributes and their immediate spatial relationships. These attribute-based checking tasks require the model to learn and recognize well-established design norms, such as typical spatial arrangements, standard family assignments, and prescribed dimensional constraints. Since these rules are prevalent and statistically significant across historical BIM datasets, the model is able to effectively generalize from pretraining and transfer this knowledge to new tasks. This leads to robust performance and low misclassification rates for these types of errors. In contrast, the third task—topological error detection—proved significantly more challenging. This task requires the model to synthesize information about the spatial coordinates, dimensions, and spatial context of multiple components to infer whether a valid topological relationship (e.g., connectivity between MEP elements and fittings) should exist. Unlike attribute errors, topological errors often arise from subtle changes in geometry or placement, posing a significant challenge and effectively distinguishing model performance. Notably, the proposed method achieves substantially higher accuracy for this challenging error type compared to non-pretrained models. This indicates that BIGNet, through node feature reconstruction, can deeply capture the intrinsic relationships within graph data.

In conclusion, the pretraining and transfer learning framework proposed in this study can effectively leverage historical data to extract and reuse design knowledge, significantly enhancing the accuracy of design checking tasks and providing precise and comprehensive information support throughout the building lifecycle. Moreover, experimental results also demonstrate that the framework can achieve high performance in transfer learning with minimal labeled data.

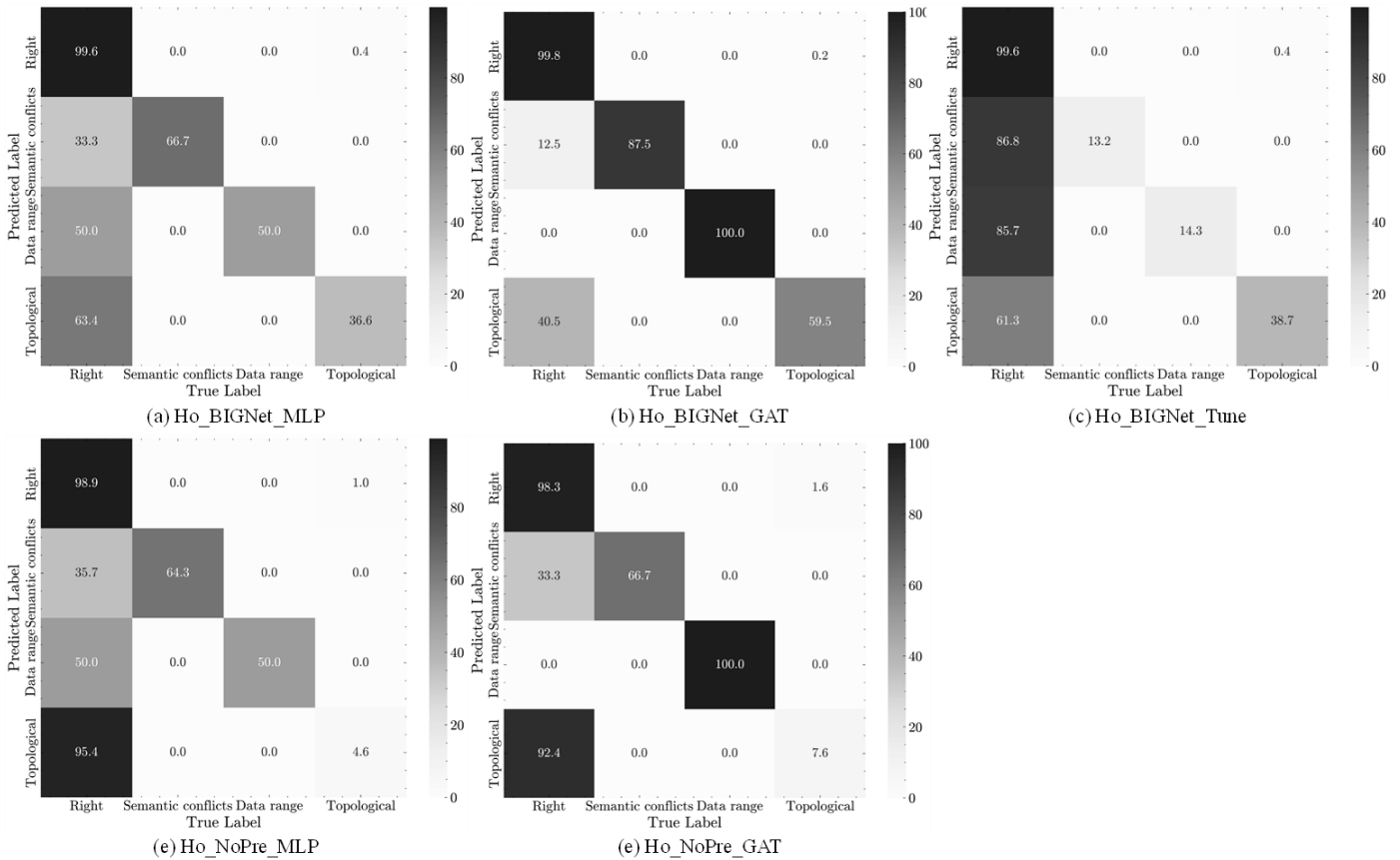

**FIGURE 19** Confusion matrix for a specific test graph under different transfer learning methods.

## 5 | DISCUSSION

This study developed BIGNet, the first pretrained graph neural network for embedding semantic, spatial, and topological data in building information models. By leveraging historical design data, BIGNet enables transfer learning for BIM-based design checking tasks with minimal labeled data, significantly improving detection accuracy while minimizing the need for extensive manual annotations. In comparison to the previous effort, this work contributes to the body of knowledge on four main levels:

(a) This study introduces a graph representation method for multi-dimensional features of BIM models and

constructs the first BIM-specific graph dataset containing nearly 1 million nodes and 3.5 million edges.

(b) Based on the above dataset, this study establishes the first pretrained model for BIM, BIGNet. Through node reconstruction, BIGNet effectively learns the semantic, spatial, and topological relationships within a graph without relying on labeled data. This significantly enhances the understanding and representation of implicit associations among BIM components, providing a robust feature foundation for downstream tasks and reducing the need for extensive manual annotation. Notably, our pretrained model is task-agnostic and can be efficiently adapted to multiple distinct BIM downstream tasks within a unified framework.

(c) The proposed method achieves an Average F1-score improvement of 59.5% (with an MLP classifier) and 72.7% (with a GAT classifier) over traditional non-pretrained methods in BIM-based design checking tasks, fully demonstrating its superiority. In practical applications, this framework allows engineers only to manually inspect and annotate a small portion of the target drawings for transfer learning, while automated detection for the remaining sections. This approach reduces human intervention while ensuring efficient integration and quality assurance across the design, construction, and operation stages of a building's lifecycle.

(d) Experimental results demonstrate that the proposed transfer learning method is well-suited for various BIM-based design checking tasks, effectively addressing common errors such as Semantic conflicts, data range errors, and topological errors. Unlike previous studies that typically focus on a single task, our approach formulates the detection problem as a node classification task (see Section 3.3.1), enabling the simultaneous detection of multiple error types within a unified framework. This unified strategy streamlines the checking process and greatly improves the versatility and efficiency of BIM model validation. Analysis of prediction outcomes further reveals that the method is particularly stringent in identifying erroneous samples, with very few incorrect nodes misclassified as correct, underscoring the reliability and practical value of the proposed approach.

(e) It is important to note that the proposed graph representation method can transform tasks such as error detection into node classification problems. This flexibility allows for the addition or removal of graph nodes and node features based on specific downstream tasks, providing excellent scalability. For example, when applying BIGNet to collision detection, classifying spatial nodes can achieve the desired outcome. Similarly, if the downstream task necessitates considering the design phase of components, this information can be incorporated into semantic node features using one-hot encoding or other methods, further enriching the model's informational depth.

(f) Beyond improving accuracy in design checking, the proposed approach holds significant potential for various advanced applications in architecture, engineering, and construction (AEC). For example, design recommendation systems can benefit from this framework by learning implicit design patterns from historical BIM models and suggesting optimal component configurations, spatial arrangements, or material choices. Besides, the integration of semantic augmentation can enhance the interoperability and usability of BIM models. By embedding richer semantic relationships and context into the graph structure, future systems could automatically enrich incomplete or simplified models, making them more informative and adaptable across design phases. In the context of design optimization, the proposed framework can facilitate multi-objective optimization processes by identifying and correcting inefficiencies or conflicts within BIM models. For example, spatial conflicts or topological inefficiencies could be detected and resolved through graph-based reasoning, enhancing both safety and performance in building systems. These capabilities significantly reduce manual intervention and improve accuracy, efficiency, and adaptability throughout the digital construction lifecycle.

Beyond the scope of the current research, future improvements can be pursued through the following approaches:

(a) Additional component features, such as time or design phase, can be incorporated into node and edge features to account for construction processes and design stage impacts, thereby expanding the applicability of downstream transfer learning tasks. For example, if applied to spatiotemporal clash detection, properties like construction priority and design phase can be extracted. Similarly, for applications in operation and maintenance, properties such as access space and access frequency can be included.

(b) The data-driven BIM design checking method proposed in this study faces limitations due to the constraints of deep learning models, making it difficult to identify all potential errors. However, it is effective in detecting implicit rules and conventional design constraints that are not explicitly specified in standards but emerge in practical design. To address this, the automated interpretation and extraction of textual rules from standards and construction plans will be integrated. By encoding these rules into a machine-readable format, the method can validate strictly defined metrics, achieving knowledge-driven "precise recognition." Additionally, combined with the proposed approach, which learns from historical design data, it enables "comprehensive coverage" of implicit rules not explicitly stated in standards. This integration can enhance the thoroughness and reliability of BIM model checking.

(c) To enable more flexible and natural heterogeneous representations, several avenues can be explored in future

work: (1) Parameter Sharing and Modular Architectures: Implement parameter sharing strategies or modular network designs that allow for efficient learning across node types, thereby reducing the risk of overfitting while maintaining expressivity; and (2) Hybrid Approaches: Develop hybrid models that combine the efficiency of homogeneous representations at the local node level with the structural richness of heterogeneous graphs at the global level, potentially leveraging hierarchical or multi-scale GNN architectures.

(d) Additionally, our work lays a foundation for future research in feature selection and dimensionality reduction within pretrained BIM-GNN models, including the integration of attention mechanisms or sparsity constraints to further improve efficiency. Besides, this study focuses on a single GNN architecture for pretraining. However, numerous scholars have proposed various improvements to GNN structures. Future work will explore diverse GNN models and pretraining methods to enhance performance and develop transfer learning models with higher recall rates.

(e) Nevertheless, we see potential in future work to integrate GNN-based feature learning with graph database infrastructures to leverage the strengths of both approaches for even greater scalability and flexibility in BIM data utilization. Specifically, graph databases offer efficient storage, querying, and real-time management of large and complex BIM datasets, supporting dynamic updates and flexible schema evolution. When combined with the powerful representation and pattern recognition capabilities of pretrained GNNs, this integration could enable seamless end-to-end pipelines—from efficient data management and retrieval to automated feature extraction and advanced analytics. Such a hybrid framework would facilitate more scalable, adaptive, and intelligent BIM applications, including real-time model checking, design recommendation, and cross-model knowledge discovery. We have also highlighted this direction in the revised discussion section to underscore its promising value for future research and industry adoption.

(f) In previous studies by the authors, structural design tasks such as shear wall layout optimization were primarily addressed using 2D architectural floor plans. Looking ahead, a promising direction lies in bridging 2D structural design generation with BIM-based modeling, ultimately enabling a seamless workflow from 2D layout optimization to 3D modeling and automated error checking.

# 6 | CONCLUSION

This research explores methods for extracting, learning, and reusing design knowledge embedded in existing BIM models. First, a scalable graph representation was proposed to encode the "semantic-spatial-topological" design features of BIM components. Subsequently, BIGNet is proposed by introducing a new message-passing mechanism to GraphMAE2 and further pretrained with a node masking strategy. To our knowledge, this is the first pretrained network tailored for graph data derived from 3D BIM models. Finally, BIGNet is evaluated in various transfer learning tasks for BIM-based design checking. The key conclusions are as follows:

(a) Compared to heterogeneous graphs, homogeneous graph representations are more effective, a 0.3m radius of local spatial relationships most enhances BIGNet's performance, and BIGNet with GAT-based feature extraction is the best approach for transfer learning.

(b) The proposed method achieves an Average F1-score improvement of 59.5% (with an MLP classifier) and 72.7% (with a GAT classifier) over traditional non-pretrained methods in BIM-based design checking tasks, providing precise and comprehensive information support throughout the building lifecycle.

(c) The proposed method is applicable to various BIM model checking tasks and demonstrates exceptional rigor in identifying erroneous samples, with very few misclassifications of incorrect nodes as correct. This further highlights the reliability of the method in practical applications.

This work demonstrates the initial feasibility and potential of a pretrained GNN framework for representing and extracting complex design knowledge from BIM data, thereby paving the way for future applications of transfer learning in BIM data utilization.

## REFERENCES

Adeli, H. (2001). Neural networks in civil engineering: 1989–2000. *Computer-Aided Civil and Infrastructure Engineering*, *16*(2), 126-142.

Adeli, H., & Park, H. S. (1995). Optimization of space structures by neural dynamics. *Neural networks*, 8(5), 769-781.

Adeli, H. & Nogay, H. S., (2021). Detection of epileptic seizure using pretrained deep convolutional neural network and transfer learning. *European neurology*, *83*(6), 602-614.

Ahmadlou, M., Adeli, H., & Adeli, A. (2012). Graph theoretical analysis of organization of functional brain networks in ADHD. *Clinical EEG and neuroscience*, *43*(1), 5-13.

Amezquita-Sanchez, J. P., Valtierra-Rodriguez, M., Aldwaik, M., & Adeli, H. (2016). Neurocomputing in civil infrastructure. *Scientia Iranica*, 23(6), 2417-2428.

Cui, G., Zhou, J., Yang, C., & Liu, Z. (2020, August). Adaptive graph encoder for attributed graph embedding. In *Proceedings of the 26th ACM SIGKDD international conference on knowledge discovery & data mining* (pp. 976-985).

Gan, V. J. (2022). BIM-based graph data model for automatic generative design of modular buildings. *Automation in Construction, 134*, 104062.

Garcia Duran, A., & Niepert, M. (2017). Learning graph representations with embedding propagation. *Advances in neural information processing system*, *30*.

Ge, K., Wang, C., Guo, Y. T., Tang, Y. S., Hu, Z. Z., & Chen, H. B. (2024). Fine-tuning vision foundation model for crack segmentation in civil infrastructures. *Construction and Building Materials*, 431, 136573.

Han, J., Lu, X., Gu, Y., Liao, W., Cai, Q., & Xue, H. (2024a).

Optimized data representation and understanding method for the intelligent design of shear wall structures. *Engineering Structures, 315*, 118500.

Han, J., Zheng, Z., Lu, X. Z., Chen, K. Y., & Lin, J. R. (2024a). Enhanced Earthquake Impact Analysis based on Social Media Texts via Large Language Model. *International Journal of Disaster Risk Reduction*, 104574.

Han, J., Lu, X. Z., & Lin, J. R. (2025, July). BIGNet: Pretrained Graph Neural Network for Embedding Semantic, Spatial, Topological Data in Building Information Models. In 42th International Symposium on Automation and Robotics in Construction (ISARC 2025) (under review). ISARC.

He, Z., Wang, Y. H., & Zhang, J. (2025). Generative AIBIM: An automatic and intelligent structural design pipeline integrating BIM and generative AI. *Information Fusion, 114*, 102654.

Hou, Z., He, Y., Cen, Y., Liu, X., Dong, Y., Kharlamov, E., & Tang, J. (2023, April). Graphmae2: A decoding-enhanced masked self-supervised graph learner. In *Proceedings of the ACM web conference 2023* (pp. 737-746).

Hou, Z., Liu, X., Cen, Y., Dong, Y., Yang, H., Wang, C., & Tang, J. (2022, August). Graphmae: Self-supervised masked graph autoencoders. In *Proceedings of the 28th ACM SIGKDD Conference on Knowledge Discovery and Data Mining* (pp. 594-604).

Hu, Y., & Castro-Lacouture, D. (2019). Clash relevance prediction based on machine learning. *Journal of computing in civil engineering, 33*(2), 04018060.

Hu, Y., Castro-Lacouture, D., Eastman, C. M., & Navathe, S. B. (2020). Automatic clash correction sequence optimization using a clash dependency network. *Automation in Construction, 115*, 103205.

Hu, Y., Xia, C., Chen, J., & Gao, X. (2023). Clash context representation and change component prediction based on graph convolutional network in MEP disciplines. *Advanced Engineering Informatics, 55*, 101896.

Hu, Z. Z., Leng, S., Lin, J. R., Li, S. W., & Xiao, Y. Q. (2022). Knowledge extraction and discovery based on BIM: a critical review and future directions. *Archives of Computational Methods in Engineering*, 29(1), 335-356.

Jang, S., & Lee, G. (2024). BIM Library transplant: bridging human expertise and artificial intelligence for customized design detailing. *Journal of Computing in Civil Engineering, 38*(2), 04024004.

Kayhani, N., McCabe, B., & Sankaran, B. (2023). Semantic-aware quality assessment of building elements using graph neural networks. *Automation in Construction, 155*, 105054.

Kipf, T. N., & Welling, M. (2016). Variational graph auto-encoders. *arXiv preprint arXiv:1611.07308*.

Leng, S., Lin, J. R., Li, S. W., & Hu, Z. Z. (2021). A data integration and simplification framework for improving site planning and building design. *IEEE Access, 9*, 148845-148861.

Li, K., Gan, V. J., Li, M., Gao, M. Y., Tiong, R. L., & Yang, Y. (2024). Automated generative design and prefabrication of precast buildings using integrated BIM and graph convolutional neural network. *Developments in the Built Environment, 18*, 100418.

Liang, J., Zhong, X., & Koh, I. (2024). BRIDGING BIM AND AI: A Graph-BIM Encoding Approach For Detailed 3D Layout Generation Using Variational Graph Autoencoder. In *International Conference on Computer-Aided Architectural Design Research in Asia* (pp. 221-230). Association for Computer-Aided Architectural Design Research in Asia.

Lin, J. R., Hu, Z. Z., Zhang, J. P., & Yu, F. Q. (2016). A natural-language-based approach to intelligent data retrieval and representation for cloud BIM. *Computer-Aided Civil and Infrastructure Engineering, 31*(1), 18-33.

Lin, J.R.*, Chen, K.Y., Pang, P., Wu, D.P., Zhang, J.P. (2024). ChatBIM: Interactive Data Retrieval and Visualization from BIM. *the 10th International Conference on Innovative Production and Construction (IPC 2024)*, 265-266. Perth, Australia.

Lin, W. Y., & Huang, Y. H. (2019). Filtering of irrelevant clashes detected by BIM software using a hybrid method of rule-based reasoning and supervised machine learning. *Applied Sciences, 9*(24), 5324.

Liu, H., Gan, V. J., Cheng, J. C., & Zhou, S. A. (2024). Automatic Fine-Grained BIM element classification using Multi-Modal deep learning (MMDL). *Advanced Engineering Informatics, 61*, 102458.

Liu, X., Zhao, J., Yu, Y., & Ji, Y. (2024). BIM-based multi-objective optimization of clash resolution: A NSGA-II approach. *Journal of Building Engineering, 89*, 109228.

Panakkat, A., & Adeli, H. (2007). Neural network models for earthquake magnitude prediction using multiple seismicity indicators. *International journal of neural systems, 17*(01), 13-33.

Rad, M. A. H., Jalaei, F., Golpour, A., Varzande, S. S. H., & Guest, G. (2021). BIM-based approach to conduct Life Cycle Cost Analysis of resilient buildings at the conceptual stage. *Automation in Construction, 123*, 103480.

Utkucu, D., Ying, H., Wang, Z., & Sacks, R. (2024). Classification of architectural and MEP BIM objects for building performance evaluation. *Advanced Engineering Informatics, 61*, 102503.

Wang, L., & Leite, F. (2013). Knowledge discovery of spatial conflict resolution philosophies in BIM-enabled MEP design coordination using data mining techniques: a proof-of-concept. In *Computing in Civil Engineering (2013)* (pp. 419-426).

Wang, Y., Zhang, L., Yu, H., & Tiong, R. L. (2022). Detecting logical relationships in mechanical, electrical, and plumbing (MEP) systems with BIM using graph matching. *Advanced Engineering Informatics, 54*, 101770.

Wang, Z., Ying, H., Sacks, R., & Borrmann, A. (2023). CBIM: A graph-based approach to enhance interoperability using semantic enrichment. *arXiv preprint arXiv:2304.11672*.

Weiss, K., Khoshgoftaar, T. M., & Wang, D. (2016). A survey of transfer learning. *Journal of Big data, 3*, 1-40.

Xiao, Y. Q., Li, S. W., & Hu, Z. Z. (2019). Automatically generating a MEP logic chain from building information models with identification rules. *Applied Sciences*, 9(11), 2204.

Zhang, Y. Y., Hu, Z. Z., Lin, J. R., & Zhang, J. P. (2021). Linking data model and formula to automate KPI calculation for building performance benchmarking. *Energy Reports*, 7, 1326-1337.

Zheng, Z., Lu, X. Z., Chen, K. Y., Zhou, Y. C., & Lin, J. R. (2022). Pretrained domain-specific language model for natural language processing tasks in the AEC domain. *Computers in Industry, 142*, 103733.

Zheng, Y., Zhang, Y., & Lin, J. (2023). BIM–based time-varying system reliability analysis for buildings and infrastructures. *Journal of Building Engineering, 76*, 106958.

Zheng, Z., Zhou, Y. C., Chen, K. Y., Lu, X. Z., She, Z. T., & Lin, J. R. (2024). A text classification-based approach for evaluating and enhancing the machine interpretability of building codes. *Engineering Applications of Artificial Intelligence*, 127, 107207.

Zou, Y., Kiviniemi, A., & Jones, S. W. (2017). A review of risk management through BIM and BIM-related technologies. *Safety science, 97*, 88-98.